\documentclass{article}

\usepackage{arxiv}

\usepackage[utf8]{inputenc} 
\usepackage[T1]{fontenc}    
\usepackage{hyperref}       
\usepackage{url}            
\usepackage{booktabs}       
\usepackage{amsfonts}       
\usepackage{nicefrac}       
\usepackage{microtype}      
\usepackage{graphicx}
\usepackage{amsmath}
\usepackage{amssymb}
\usepackage{multirow}
\usepackage{wrapfig}
\usepackage{subcaption}
\usepackage[table]{xcolor}
\usepackage{xspace}
\usepackage[capitalise,noabbrev]{cleveref}

\definecolor{rowgray}{gray}{0.85}
\definecolor{rowblue}{RGB}{218,232,252}

\newcommand{\eg}{\textit{e.g.}\xspace}

\newcommand{\cf}{\textit{cf.}\xspace}

\graphicspath{ {./} }

\title{The Model Knows Which Tokens Matter:\\Automatic Token Selection via Noise Gating}

\author{
  Landi He \\
  Shenzhen University of Advanced Technology \\
  \And
  Xiaoyu Yang \\
  Shenzhen University of Advanced Technology \\
  \And
  Lijian Xu  \thanks{Corresponding author} \\
  Shenzhen University of Advanced Technology \\
  \texttt{xulijian@suat-sz.edu.cn}
} 

\begin{document}
\maketitle

\begin{abstract}
Visual tokens dominate inference cost in vision-language models (VLMs), yet many carry redundant information.
Existing pruning methods alleviate this but typically rely on attention magnitude or similarity scores.
We reformulate visual token pruning as capacity constrained communication: given a fixed budget $K$, the model must allocate limited bandwidth to maximally preserve visual information.
We propose \textbf{AutoSelect}, which attaches a lightweight Scorer and Denoiser to a frozen VLM and trains with only the standard next token prediction loss, without auxiliary objectives or extra annotations.
During training, a variance preserving noise gate modulates each token's information flow according to its predicted importance so that gradients propagate through all tokens; a diagonal attention Denoiser then recovers the perturbed representations.
At inference, only the Scorer and a hard top-$K$ selection remain, adding negligible latency.
On ten VLM benchmarks, AutoSelect retains 96.5\% of full model accuracy while accelerating LLM prefill by 2.85$\times$ with only 0.69\,ms overhead, and transfers to different VLM backbones without architecture-specific tuning. Code is available at \url{https://github.com/MedHK23/AutoSelect}.
\end{abstract}

\section{Introduction}
\label{sec:intro}

Vision-language models (VLMs) that couple a pretrained visual encoder with a large language model (LLM) have become the prevailing paradigm for visual question answering, multimodal dialogue, and cross-modal reasoning. In the standard pipeline, patch or grid features from the encoder are projected into the LLM's embedding space and typically prepended as visual tokens for autoregressive decoding, as seen in representative models like BLIP-2, InstructBLIP, and LLaVA~\cite{li2023blip2,dai2023instructblip,liu2023visual_instruction_tuning}. However, as these models are increasingly applied to high-resolution images and multi-image or video scenarios, the resulting surge in visual tokens creates a severe computational bottleneck. Due to the quadratic scaling of self-attention with respect to sequence length, these abundant visual tokens quickly dominate both inference computation and memory.

Empirical studies reveal substantial redundancy among visual tokens. Attention distributions are typically highly concentrated, with only a small subset of tokens receiving significant attention while a large fraction exhibits near-zero attention across layers~\cite{chen2024fastv,yang2025visionzip}. Nevertheless, subsequent layers still allocate full self-attention computation to all tokens, including those with negligible contribution to the final prediction. These observations suggest that substantial computational redundancy exists in current VLM pipelines.

Existing pruning methods tackle this challenge from multiple directions, including inference-time token and key-value (KV) cache optimization~\cite{yang2025topv}, layer-wise budget search~\cite{zhao2025_gsearch}, and instruction-aware or cross-modal selection strategies~\cite{huang2024ivtp,cao2024madtp,ye2025_atp_llava}. While these approaches achieve useful speed--accuracy trade-offs, their selection criteria are typically based on local proxy signals such as attention magnitude, similarity scores, or predefined pruning schedules. Consequently, pruning is primarily formulated as identifying and discarding ``less important'' tokens. This perspective overlooks a more fundamental question: given a fixed computational budget, how should representational capacity be globally allocated across visual tokens to maximize downstream reasoning performance?

\setlength{\intextsep}{0pt}%
\setlength{\columnsep}{8pt}%
\begin{wrapfigure}{r}{0.5\linewidth}
  \centering
  \includegraphics[width=\linewidth]{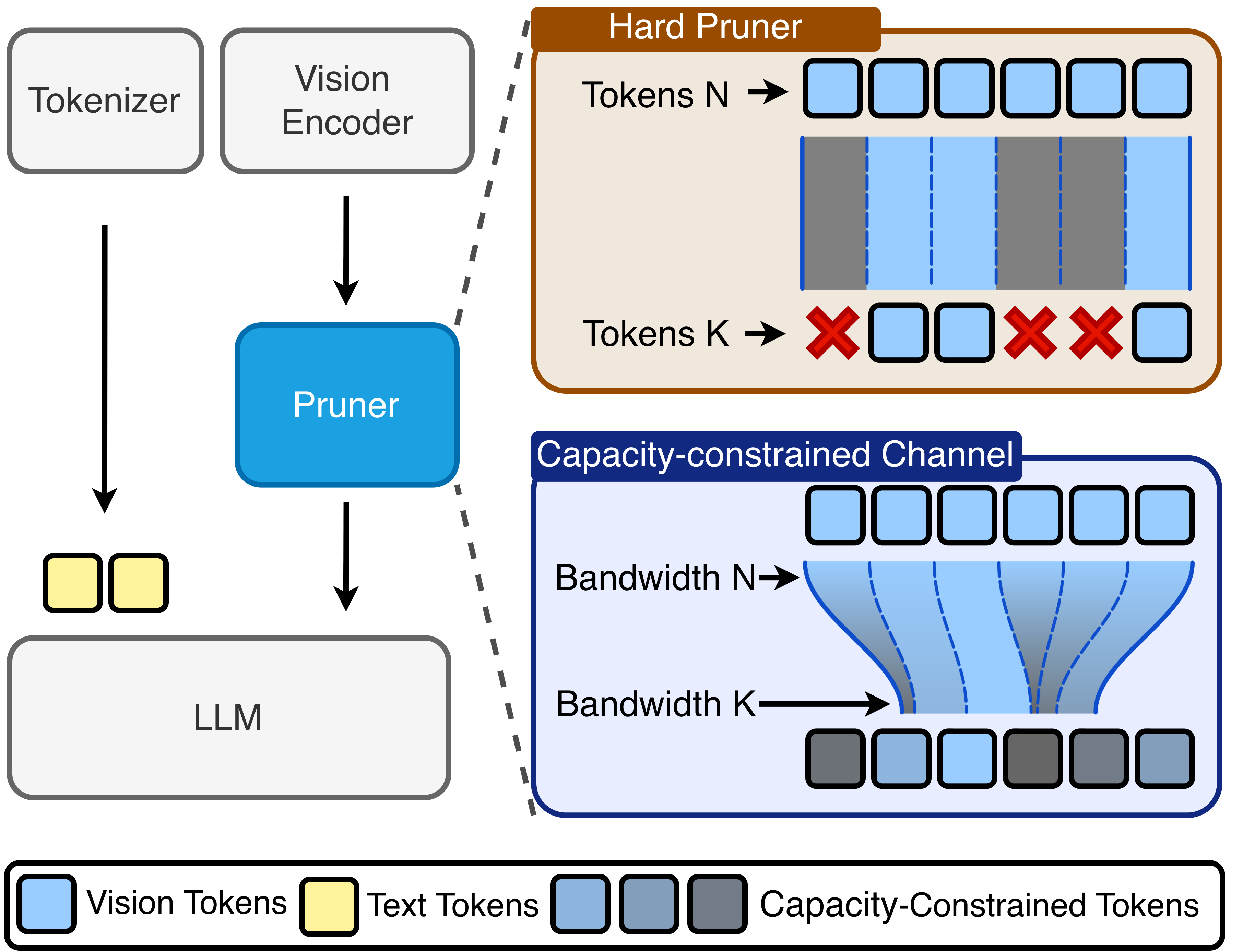}
  \caption{
    \textbf{Two views of visual token pruning.}
    \textit{(Top)} Hard pruning directly discards a subset of tokens.
    \textit{(Bottom)} Our capacity-constrained formulation retains all tokens but limits the total information throughput to the same budget through a bandwidth-limited channel.
  }
  \label{fig:overview}
\end{wrapfigure}

We approach this problem by recasting visual token pruning as \textbf{capacity-constrained representation learning}. As illustrated in \cref{fig:overview}, a pruning module sits between the visual encoder and the LLM. Existing methods treat this module as a hard filter that irreversibly discards tokens, whereas we model the same encoder--LLM interface as a bandwidth-limited channel in which a fixed token budget~$K$ constrains the total information capacity rather than the token count. During training, no tokens are removed; instead, the effective information throughput of each token is continuously modulated by its importance score, so that a budget of $K$ tokens worth of capacity is allocated to the most informative content. Under this formulation, the objective shifts from identifying dispensable tokens to constructing a compact representation that maximizes useful information within the prescribed capacity.

We introduce two lightweight plug-in modules, a Scorer and Denoiser, while keeping the pretrained VLM entirely frozen. During training, no tokens are discarded; importance scores from the Scorer modulate information flow through a continuous bottleneck, attenuating low-scoring tokens under a fixed capacity constraint. This reformulates discrete pruning as a differentiable capacity allocation objective optimized end-to-end. A Denoiser then remaps noise-perturbed tokens back into the distribution expected by the frozen LLM, operating independently on each token to prevent information leakage. At inference, the continuous bottleneck is replaced by hard top-$K$ selection: only the highest-scoring tokens are forwarded to the LLM, yielding computational savings without modifying the base architecture.

Our contributions are threefold:

1) We reformulate visual token pruning as \emph{capacity-constrained representation learning}, modeling the visual encoder--LLM interface as a bandwidth-limited channel. The resulting framework is optimized with the standard next-token prediction loss as its sole training objective, without auxiliary losses, external annotations, or modifications to the base VLM.

2) We introduce a variance-preserving noise gate that replaces the binary keep-or-discard decision of conventional pruning with continuous per-token information capacity modulation. Paired with a Soft Top-$K$ operator and temperature annealing, this mechanism provides full gradient flow during training while converging to hard Top-$K$ selection at inference.

3) AutoSelect retains 96.5\% of full-model performance at 88.9\% pruning on LLaVA-1.5-7B with only 0.69\,ms pruning module overhead, and generalizes consistently to the higher-resolution LLaVA-NEXT and the architecturally distinct Qwen2.5-VL.

\section{Related Work}
\label{sec:related-work}

\subsection{VLMs and Efficient Paradigms}
Large vision-language models (VLMs) have rapidly progressed by coupling visual perception with the generative and reasoning capabilities of large language models (LLMs)~\cite{shen2025vlm,kim2024image,xu2021vlm,zeng2023x}.
Most contemporary open VLMs adopt an ``encode-and-project'' paradigm, where image patches are encoded into visual tokens and mapped into the LLM token space via learned projectors, cross-attention, or query-based resamplers~\cite{liu2023visual_instruction_tuning,li2023blip2,flamingo}.
Instruction tuning and dialogue-style supervision further align these models to follow human prompts and generalize across vision-language tasks~\cite{liu2023visual_instruction_tuning,dai2023instructblip}. Beyond encoder-based pipelines, EVE~\cite{eve} explores encoder-free unified-decoder training to improve architectural flexibility and efficiency when mixing vision and language streams.

As VLMs scale to richer visual inputs, including high-resolution images, arbitrary aspect ratios, multi-image interleaving, and video, the number of visual tokens increases rapidly. This growth substantially amplifies quadratic attention complexity and KV-cache overhead in the LLM, as demonstrated by LLaVA-NeXT~\cite{llava-next-interleave}, LLaVA-OneVision~\cite{llava-onevision}, and Qwen2-VL~\cite{qwen2-vl}. To enhance fine-grained perception without prohibitive token costs, MRA~\cite{mra} adopts multi-resolution pathways to selectively attend to fine-grained tokens, Mini-Gemini~\cite{li2025mini} employs a dual visual encoder where low-resolution tokens query high-resolution patches via patch-level information mining, and PuMer~\cite{cao2023pumer} progressively reduces visual and textual tokens through text-informed pruning and modality-aware merging. While architectural and operator-level optimizations (\eg, windowed attention~\cite{swin}, fused kernels~\cite{flashattention,flashattention2,flashattention3}) improve overall throughput, the visual token count entering the LLM remains a dominant cost factor, motivating dedicated token reduction techniques.
Recent advances in efficient visual representation learning, including occlusion-based contrastive learning~\cite{yang2025one,feng2026efficient}, self-adaptive token bases~\cite{young2026fewer}, and robust spatial-concept alignment~\cite{young2026scalar}, have shown that visual features can be substantially compressed without sacrificing discriminative power. These principles have also proven effective in domain-specific tasks such as medical image interpretation~\cite{xu2023learning,yang2024segmentation,yang2023geometry}, pathology token compression~\cite{chen2026tc,wu2026towards}, multimodal medical foundation models~\cite{xu2024medvilam,xu2024foundation} and vision-centric long-context compression~\cite{gao2026zerosense}, further motivating our capacity-constrained formulation for token pruning.

\subsection{Visual Token Reduction}
Token reduction in transformer models is commonly achieved through token pruning or token merging, and has been widely studied in vision transformers to reduce computation while preserving performance~\cite{tome,dynamic-token-pruning,zerotprune}.
In VLMs, the efficiency bottleneck often shifts to the LLM prefill stage, where long visual-token prefixes substantially increase self-attention cost and KV storage~\cite{cao2025survey}. Therefore, reducing prefix visual tokens becomes a direct lever for lowering latency and memory.
A representative line of work performs plug-and-play, training-free token pruning right after the vision encoder.
PruMerge~\cite{prumerge} and FasterVLM~\cite{fastervlm} use vision-side statistics such as [CLS]-to-patch attention to score token saliency. Other training-free approaches emphasize representational coverage: DivPrune~\cite{divprune}, DART~\cite{dart}, and Feather~\cite{feather-throttle} select tokens to maximize diversity or to minimize duplication among the retained set, which can be more robust than pure importance ranking at high pruning ratios.

Beyond one-shot pruning, FitPrune~\cite{fitprune}, VTW~\cite{vtw}, and Balanced Token Pruning~\cite{balanced-token-pruning} introduce calibration-based or staged schedules to balance local feature preservation and downstream effects, typically using a small calibration set or lightweight statistics. In contrast to post-encoder pruning, FastV~\cite{chen2024fastv} and SparseVLM~\cite{sparsevlm} prune inside the LLM stack to exploit deeper semantic mixing, trading earlier compute for potentially better task retention. Recent analyses by FasterVLM~\cite{fastervlm}, DART~\cite{dart}, and \cite{token-pruning-right-problem} highlight that pruning signal and evaluation protocol are critical: cross-attention can be a noisy proxy for token importance, and naive baselines may be surprisingly competitive. From an engineering standpoint, strategies requiring dynamic token selection inside transformer blocks can complicate integration with optimized kernels like FlashAttention~\cite{flashattention,flashattention2,flashattention3}, whereas prefix-only pruning is often easier to deploy~\cite{dart}.

The pruning location also interacts with the accuracy--speed trade-off. Late pruning leverages higher-level semantics but reduces achievable end-to-end speedup, while early pruning maximizes savings but risks discarding fine-grained evidence needed for text-rich or localization-sensitive queries~\cite{chen2024fastv,vtw,flexattention,fastvlm}.
Finally, [CLS]-based scoring methods such as PruMerge~\cite{prumerge} and FasterVLM~\cite{fastervlm} implicitly assume the backbone provides such a token, which may not hold for modern architectures like Qwen2.5-VL~\cite{bai2025qwen25vltechnicalreport}, whose vision encoder lacks a [CLS] token.

Several recent methods learn token selection end-to-end rather than relying on fixed heuristics.
GlimpsePrune~\cite{zeng2025glimpse} trains a visual importance predictor before answer generation; however, its optimization relies heavily on bounding box annotations, tying pruning to a localization surrogate and potentially limiting generalization to open-vocabulary tasks.
ATP-LLaVA~\cite{ye2025_atp_llava} incorporates adaptive modules between LLM layers to predict instance-wise retention ratios. Although it uses differentiable approximations to bypass hard selection, it still requires complex auxiliary losses to enforce pruning budgets.
Taking a different direction, p-MoD~\cite{zhang2025pmod} applies Mixture-of-Depths routing to vision tokens. Ensuring convergence requires joint fine-tuning of the entire LLM backbone, an intrusive process risking disruption of pre-trained language priors.
Together with the heuristic limitations discussed above, these observations motivate a pruning formulation that is data-driven yet non-intrusive to the base model.

\section{Methodology}

\begin{figure}[t]
  \centering
  \includegraphics[width=0.8\linewidth]{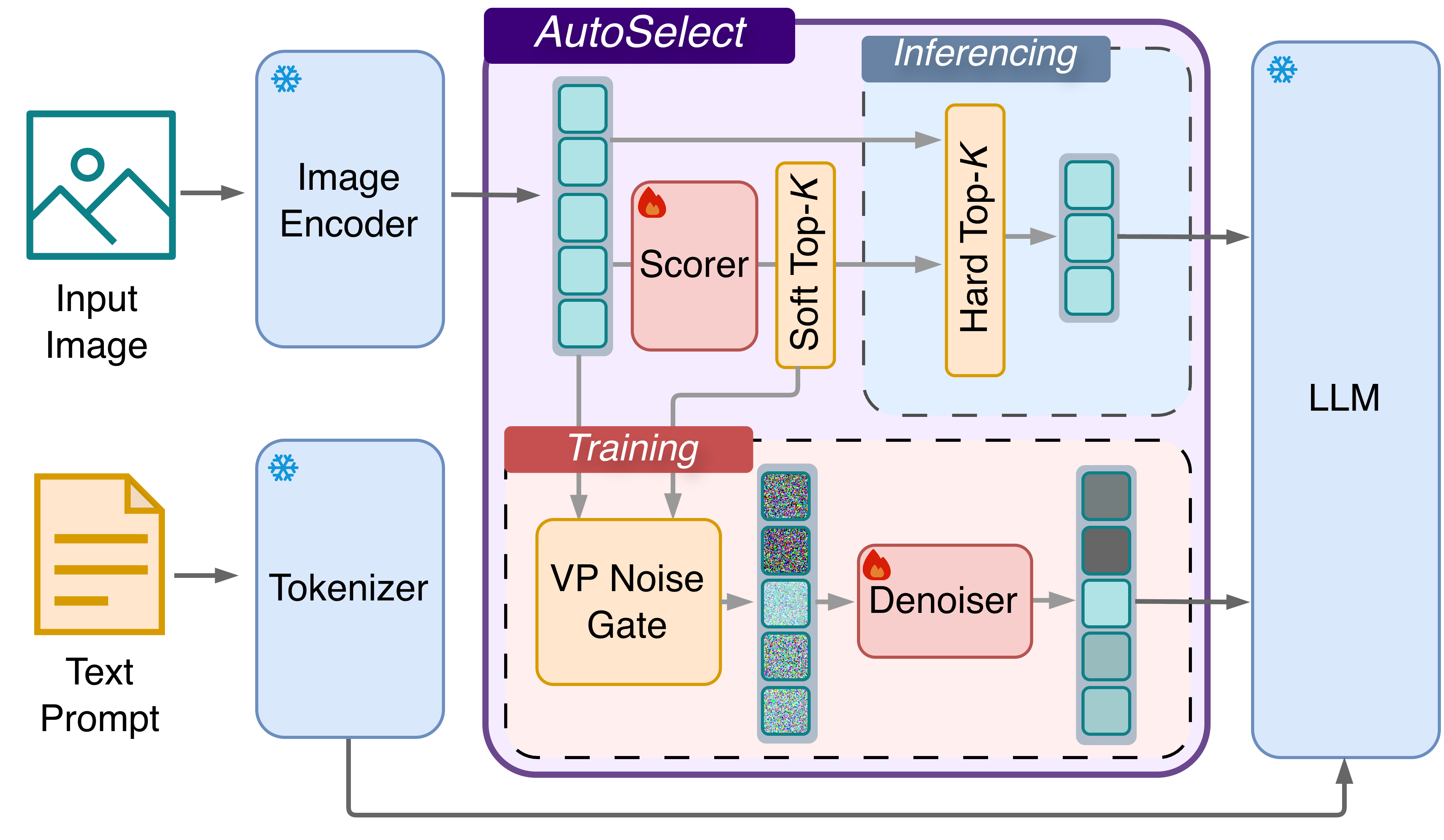}
  \caption{
    \textbf{Overview of the AutoSelect framework.}
    Visual tokens from the frozen Image Encoder pass through a learnable Scorer that assigns per-token importance scores.
    These scores are polarized by the differentiable \emph{Soft Top-$K$} operator under a fixed bandwidth budget~$K$.
    \textit{During training} (lower path), a VP Noise Gate injects variance-preserving noise into each token in inverse proportion to its score; the Denoiser then maps the perturbed sequence back toward the LLM's expected input space.
    \textit{At inference} (upper path), the Denoiser and noise injection are discarded: Hard Top-$K$ retains the $K$ highest-scoring tokens with their original position indices.
    All base VLM parameters, including the Image Encoder, modality projector, and LLM, remain frozen.
  }
  \label{fig:framework_overview}
\end{figure}

\subsection{Overview of the Framework}

Given an input image $\mathbf{I}$, a frozen vision encoder $\mathcal{E}_v$ produces $N$ visual token embeddings $\mathbf{X}^v = \mathcal{E}_v(\mathbf{I}) \in \mathbb{R}^{N \times d_v}$, transformed by a projector $\mathbf{P}_{v\to\ell}$ and concatenated with text embeddings $\mathcal{E}_t(\mathbf{T})$ before being fed to the frozen LLM $\mathcal{L}$.
Since the vision encoder accounts for less than 5\% of total inference cost while the LLM dominates, we place our pruning module after the encoder and before the projector to maximize computational savings.

As illustrated in \cref{fig:framework_overview}, our framework introduces two lightweight modules, a \textbf{Scorer} $\mathcal{S}(\cdot)$ and a \textbf{Denoiser} $\mathcal{D}(\cdot)$, which are jointly trained and placed between the encoder and the projector, while all original VLM parameters remain frozen.
The entire framework is optimized end-to-end using the standard language modeling objective, without requiring auxiliary losses or hand-crafted features.

\paragraph{Training phase.}
The Scorer assigns per-token importance scores that are polarized by a differentiable \emph{Soft Top-$K$} operator under a fixed bandwidth~$K$.
Rather than removing tokens, we inject variance-preserving (VP) noise with magnitude inversely proportional to each token's polarized score, keeping the sequence length at~$N$.
The perturbed sequence is processed by the Denoiser and forwarded through the frozen projector and LLM.
Only the Scorer parameters~$\theta$ and Denoiser parameters~$\phi$ are optimized via the negative log-likelihood (NLL) loss for next-token prediction:
\begin{equation}
  \min_{\theta,\phi}\;
  \mathcal{J}_{\mathrm{NLL}}\!\left(
    \mathcal{L}\!\left([\mathcal{D}_\phi(\tilde{\mathbf{X}}^{v})\,\mathbf{P}_{v\to\ell}\,;\,\mathcal{E}_t(\mathbf{T})]\right),\;
    \{y^{*}_t\}
  \right),
  \label{eq:training_obj}
\end{equation}
where $\tilde{\mathbf{X}}^{v}$ denotes the noise-gated visual sequence (\cref{eq:vp_noise}) and $\{y^{*}_t\}$ the ground-truth targets.

\paragraph{Inference phase.}
At inference time, the Denoiser and noise injection are removed entirely.
The Scorer produces importance scores, and a standard hard top-$K$ operation selects the $K$ highest-scoring tokens:
\begin{equation}
  \hat{\mathbf{X}}^v = \operatorname{Top\text{-}K}\!\left(\mathbf{X}^v,\; \mathcal{S}(\mathbf{X}^v),\; K\right) \in \mathbb{R}^{K \times d_v},
\end{equation}
where the selected tokens retain their original position indices rather than being re-indexed sequentially.
This design ensures that the rotary position embeddings (RoPE) within the LLM correctly encode each retained token's spatial location in the image grid.
Because the Scorer operates only on visual features, it is text-agnostic: importance scores do not depend on the language prompt and can be reused across dialogue turns without re-evaluation.

\subsection{Learnable Token Scorer with Soft Top-\textit{K} Selection}

The Scorer $\mathcal{S}(\cdot)$ comprises $L$ Transformer encoder blocks followed by a linear projection that maps each of the $N$ visual tokens to a scalar importance score:
\begin{equation}
  \mathbf{s} = \mathcal{S}(\mathbf{X}^v) \in \mathbb{R}^{N}.
\end{equation}
A standard hard top-$K$ is piecewise constant and produces zero gradients almost everywhere, preventing effective Scorer training.
We therefore employ the differentiable Soft Top-$K$ operator~$\Phi_K$~\cite{softtopk-su}.
Raw scores are first z-score normalized for numerical stability, then mapped through~$\Phi_K$:
\begin{equation}
  \boldsymbol{\alpha} = \Phi_K\!\left(\hat{\mathbf{s}} \,/\, \tau \right) \in [0,1]^{N},
  \quad \text{with}\quad
  \sum_{i=1}^{N} \alpha_i \approx K,
  \label{eq:soft_topk}
\end{equation}
where $\hat{\mathbf{s}}$ denotes the normalized scores and $\tau > 0$ is a temperature parameter.
Conceptually, $\Phi_K$ is closer to softmax than to hard top-$K$: both are smooth, temperature-scaled maps from $\mathbb{R}^N$ to $[0,1]^N$.
The distinction is in the normalizing constraint---softmax imposes $\sum_i \alpha_i = 1$, whereas $\Phi_K$ fixes $\sum_i \alpha_i = K$, turning the operator into a budget-constrained soft assignment.
A data-dependent threshold further separates $\Phi_K$ from softmax: scores are driven toward $0$ or~$1$, so the output is bimodal rather than spread across all tokens.
Because the budget is fixed, the Scorer learns \emph{which} tokens to retain, not \emph{how many}.
We anneal $\tau$ from $\tau_{\mathrm{start}}$ to $\tau_{\mathrm{end}}$ on a cosine schedule; at large $\tau$ the scores are diffuse, and as $\tau\!\to\!0$ they collapse to the binary mask used at inference.

\subsection{Capacity-Constrained Gating via Noise Injection}

Given the polarized importance scores $\boldsymbol{\alpha}$ from the Soft Top-$K$ operator, we now describe how they are used to modulate the information content of each token during training.
A na\"{i}ve approach would be to directly remove low-scoring tokens; however, this would reduce the sequence length and introduce a non-differentiable discontinuity that disrupts gradient flow.
Instead, we keep all $N$ tokens but impose token-wise capacity constraints through noise injection: the effective information that each token can transmit to the downstream LLM is reduced in proportion to its importance score.

Concretely, we adopt a variance-preserving (VP) noise injection scheme.
For the $i$-th visual token $\mathbf{x}^v_i \in \mathbb{R}^{d_v}$, the gated representation is computed as:
\begin{equation}
  \tilde{\mathbf{x}}_i
  = \sqrt{\alpha_i}\;\mathbf{x}^v_i
  + \sqrt{1 - \alpha_i}\;\boldsymbol{\epsilon}_i,
  \quad
  \boldsymbol{\epsilon}_i \sim \mathcal{N}(\mathbf{0},\, \mathbf{I}),
  \label{eq:vp_noise}
\end{equation}
where $\alpha_i \in [0,1]$ is the polarized score from \cref{eq:soft_topk} and $\mathbf{I} \in \mathbb{R}^{d_v \times d_v}$ is the identity matrix.
This formulation admits a direct information-theoretic interpretation.
When $\alpha_i \to 1$ (high importance), the noise component vanishes and the original token is preserved; when $\alpha_i \to 0$ (low importance), the signal is replaced by isotropic Gaussian noise.
Intermediate values interpolate between these extremes, providing a differentiable proxy for discrete token removal.

Because the vision encoder output passes through layer normalization, the coefficients $\sqrt{\alpha_i}$ and $\sqrt{1 - \alpha_i}$ ensure $\mathrm{Var}(\tilde{\mathbf{x}}_i) \approx \mathrm{Var}(\mathbf{x}^v_i)$, keeping the feature scale stable and preventing distribution shifts for the frozen LLM.
The Scorer therefore determines the noise level of each token, and thus its effective capacity, according to its estimated importance.
We empirically verify that VP noise gating approximates the information restriction of hard Top-K pruning; quantitative and qualitative comparisons are shown in \cref{fig:vp_validation} (\cref{sec:ablation}).

\subsection{Lightweight Denoiser with Diagonal Attention}
Although the VP formulation preserves marginal variance, it shifts token distributions in feature space, particularly for low-importance tokens approaching isotropic Gaussian noise.
To map the perturbed sequence back toward the input distribution expected by the LLM, we introduce a lightweight Denoiser $\mathcal{D}(\cdot)$ consisting of a single Transformer encoder block.

If standard global self-attention were used, high-importance tokens could leak information to low-importance ones, undermining the capacity constraints imposed by noise injection.
We therefore adopt diagonal attention: an identity attention mask restricts each token to attend only to itself, so the self-attention layer degenerates into a per-token nonlinear transformation through the value projection and feed-forward network independently.
This prevents cross-token information leakage while enabling a learned per-token mapping from the noise-perturbed space to the LLM-compatible manifold.
The Denoiser is used only during training; at inference no noise is injected and only the top-$K$ tokens are retained, so the Denoiser adds zero overhead.

\section{Experiments}
\label{sec:experiments}

\subsection{Experimental Setup}
\label{sec:exp_setup}

The Scorer and Denoiser are jointly trained on ImageNet-1K~\cite{deng2009imagenet} captioning data from ImageNet-1K-VL-Enriched~\cite{visual_layer_imagenet1k_vl_enriched}, adding ${\sim}$84M trainable parameters while all base VLM weights remain frozen.
We evaluate on three architectures: LLaVA-v1.5-7B~\cite{liu2023visual_instruction_tuning}, LLaVA-NeXT-7B~\cite{liu2024llavanext}, and Qwen2.5-VL-7B~\cite{bai2025qwen25vltechnicalreport}.
Full training hyperparameters and implementation details are provided in the supplementary material.

\paragraph{Benchmarks.}
We evaluate on ten standard VLM benchmarks:
GQA~\cite{hudson2019gqa},
MMBench~\cite{liu2024mmbench},
MMBench-CN~\cite{liu2024mmbench},
MME~\cite{fu2023mme},
POPE~\cite{pope},
ScienceQA-IMG~\cite{lu2022learn},
VQAv2~\cite{goyal2017making},
TextVQA~\cite{singh2019towards},
SEED-Bench~\cite{li2023seed}, and
VizWiz~\cite{bigham2010vizwiz}.
This set follows the evaluation protocol used by DivPrune~\cite{divprune}, SparseVLM~\cite{sparsevlm}, and HoloV~\cite{holov}, allowing direct comparison; detailed benchmark descriptions are provided in the supplementary material.
In all tables, the ``Avg.'' column reports \emph{average performance retention}: for each benchmark, we compute the ratio of the pruned model's score to the unpruned upper bound, then average these ratios.

\paragraph{Baselines.}
We organize baselines by where pruning occurs, as this directly governs the accuracy--efficiency trade-off.
1) \textbf{Pre-LLM pruning} (same location as AutoSelect): tokens are selected or merged before the LLM, so every LLM layer processes only $K$ visual tokens.
Methods: ToMe~\cite{tome}, HiRED~\cite{hired}, DivPrune~\cite{divprune}, VisionZip~\cite{yang2025visionzip}, HoloV~\cite{holov}, and PRUNESID~\cite{fangprune}.
2) \textbf{In-LLM pruning}: tokens are pruned inside the LLM (typically at layer~2--4); the first few LLM layers still see all $N$ tokens, which retains more information at the cost of less end-to-end speedup.
Methods: FastV~\cite{chen2024fastv}, SparseVLM~\cite{sparsevlm}, DART~\cite{dart}, and PDrop~\cite{pyramiddrop}.
We report the pruning location alongside accuracy in every table so that readers can account for this structural difference when comparing results (\cf~\cref{sec:efficiency}).

\subsection{Main Benchmark Results Across Models}
\label{sec:main_results}

\begin{table}[t]
  \centering
  \caption{
    \textbf{Results on LLaVA-1.5-7B under different token budgets.}
    ``Loc.'' denotes pruning location (Pre = before LLM; L$a$--L$b$ = from LLM layer $a$ to $b$).
    Upper bound uses all 576 visual tokens; remaining blocks report 192/128/64 retained tokens
    (66.7\%/77.8\%/88.9\% pruning).
    ``Avg.'' is the mean ratio of each benchmark score to its upper bound.
    \textbf{Formatting:} within each Retain-$K$ block, the best Avg.\ is colored \textcolor{red}{red} and the second best is colored \textcolor{blue}{blue}; our method is highlighted in~\colorbox{rowblue}{blue}.
  }
  \label{tab:main_results}
  \scriptsize
  \setlength{\tabcolsep}{3pt}
  \resizebox{\linewidth}{!}{%
  \begin{tabular}{lccccccccccccc}
    \toprule
    Method & Loc. & GQA & MMB & MMB$_{\text{CN}}$ & MME & POPE & SQA & VQA$^{\text{v2}}$ & VQA$^{\text{Text}}$ & SEED & VizWiz & Avg. \\
    \midrule
    \rowcolor{rowgray} \multicolumn{13}{c}{\textit{Upper Bound, 576 Tokens (100\%)}} \\
    Vanilla & - & 61.9 & 64.7 & 58.1 & 1862 & 85.9 & 69.5 & 78.5 & 58.2 & 60.5 & 54.3 & 100\% \\
    \midrule
    \rowcolor{rowgray} \multicolumn{13}{c}{\textit{Retain 192 Tokens (66.7\% pruning)}} \\
    ToMe~{\scriptsize\textcolor{gray}{(ICLR'23)}} & Pre & 54.3 & 60.5 & - & 1563 & 72.4 & 65.2 & 68.0 & 52.1 & - & - & 88.5\% \\
    FastV~{\scriptsize\textcolor{gray}{(ECCV'24)}} & L2--L32 & 52.7 & 61.2 & 53.5 & 1612 & 64.8 & 67.3 & 67.1 & 52.5 & 57.1 & 50.8 & 89.4\% \\
    SparseVLM~{\scriptsize\textcolor{gray}{(ICML'25)}} & L1--L32 & 57.6 & 62.5 & 58.6 & 1721 & 83.6 & 69.1 & 75.6 & 56.1 & 55.8 & 50.5 & 95.8\% \\
    DART~{\scriptsize\textcolor{gray}{(EMNLP'25)}} & L2 & 60.0 & 63.6 & 57.1 & 1856 & 82.8 & 69.8 & 76.7 & 57.4 & 51.5 & 54.9 & 97.3\% \\
    HiRED~{\scriptsize\textcolor{gray}{(AAAI'25)}} & Pre & 58.7 & 62.8 & 54.7 & 1737 & 82.8 & 68.4 & 74.9 & 47.4 & - & 50.1 & 93.7\% \\
    PDrop~{\scriptsize\textcolor{gray}{(CVPR'25)}} & L8--L24 & 57.3 & 63.6 & 56.8 & 1797 & 82.3 & 69.2 & 75.1 & 56.5 & 54.7 & - & 96.0\% \\
    DivPrune~{\scriptsize\textcolor{gray}{(CVPR'25)}} & Pre & 60.0 & 62.3 & - & 1752 & 87.0 & 68.7 & 75.5 & 56.4 & 58.6 & 55.6 & 97.8\% \\
    VisionZip~{\scriptsize\textcolor{gray}{(CVPR'25)}} & Pre & 59.3 & 63.0 & 57.3 & 1783 & 85.3 & 68.9 & 76.8 & 57.3 & 58.5 & 54.1 & 97.9\% \\
    HoloV~{\scriptsize\textcolor{gray}{(NeurIPS'25)}} & Pre & 59.0 & 65.4 & 58.0 & 1820 & 85.6 & 69.8 & 76.7 & 57.4 & - & 50.9 & 98.2\% \\
    PRUNESID~{\scriptsize\textcolor{gray}{(ICLR'26)}} & Pre & 60.1 & 63.7 & - & 1791 & 86.9 & 68.5 & 76.8 & 56.7 & 59.0 & 55.4 & \textcolor{red}{98.5}\% \\
    \rowcolor{rowblue} \textbf{AutoSelect (Ours)} & Pre & 57.8 & 63.4 & 57.5 & 1791 & 86.5 & 70.2 & 76.6 & 55.3 & 59.2 & 55.9 & \textcolor{blue}{98.2}\% \\
    \midrule
    \rowcolor{rowgray} \multicolumn{13}{c}{\textit{Retain 128 Tokens (77.8\% pruning)}} \\
    ToMe~{\scriptsize\textcolor{gray}{(ICLR'23)}} & Pre & 52.4 & 53.3 & - & 1343 & 62.8 & 59.6 & 63.0 & 49.1 & - & - & 80.4\% \\
    FastV~{\scriptsize\textcolor{gray}{(ECCV'24)}} & L2--L32 & 49.6 & 56.1 & 55.9 & 1490 & 59.6 & 60.2 & 61.8 & 50.6 & 55.9 & 51.3 & 85.2\% \\
    SparseVLM~{\scriptsize\textcolor{gray}{(ICML'25)}} & L1--L32 & 56.0 & 60.0 & 51.1 & 1696 & 80.5 & 67.1 & 73.8 & 54.9 & 53.4 & 51.4 & 92.4\% \\
    DART~{\scriptsize\textcolor{gray}{(EMNLP'25)}} & L2 & 58.7 & 63.2 & 57.3 & 1840 & 80.1 & 69.1 & 75.9 & 56.4 & 50.5 & 55.3 & 96.2\% \\
    HiRED~{\scriptsize\textcolor{gray}{(AAAI'25)}} & Pre & 57.2 & 61.5 & 53.6 & 1710 & 79.8 & 68.1 & 73.4 & 46.1 & - & 51.3 & 92.2\% \\
    PDrop~{\scriptsize\textcolor{gray}{(CVPR'25)}} & L8--L24 & 57.1 & 61.6 & 56.6 & 1761 & 82.3 & 68.4 & 72.9 & 56.6 & 53.3 & - & 94.7\% \\
    DivPrune~{\scriptsize\textcolor{gray}{(CVPR'25)}} & Pre & 59.2 & 62.3 & 54.8 & 1752 & 86.9 & 69.0 & 74.7 & 56.0 & 57.1 & 55.6 & 96.9\% \\
    VisionZip~{\scriptsize\textcolor{gray}{(CVPR'25)}} & Pre & 57.6 & 62.0 & 56.7 & 1762 & 83.2 & 68.9 & 75.6 & 56.8 & 57.1 & 54.5 & 96.6\% \\
    HoloV~{\scriptsize\textcolor{gray}{(NeurIPS'25)}} & Pre & 57.7 & 63.9 & 56.5 & 1802 & 82.8 & 69.8 & 75.5 & 56.8 & - & 51.5 & 96.8\% \\
    PRUNESID~{\scriptsize\textcolor{gray}{(ICLR'26)}} & Pre & 58.8 & 62.1 & - & 1749 & 86.5 & 68.3 & 75.3 & 54.7 & 57.8 & 55.8 & \textcolor{blue}{96.9}\% \\
    \rowcolor{rowblue} \textbf{AutoSelect (Ours)} & Pre & 57.5 & 62.9 & 57.4 & 1765 & 85.7 & 70.2 & 76.1 & 54.9 & 58.4 & 56.1 & \textcolor{red}{97.6}\% \\
    \midrule
    \rowcolor{rowgray} \multicolumn{13}{c}{\textit{Retain 64 Tokens (88.9\% pruning)}} \\
    ToMe~{\scriptsize\textcolor{gray}{(ICLR'23)}} & Pre & 48.6 & 43.7 & - & 1138 & 52.5 & 50.0 & 57.1 & 45.3 & - & - & 70.1\% \\
    FastV~{\scriptsize\textcolor{gray}{(ECCV'24)}} & L2--L32 & 46.1 & 48.0 & 52.7 & 1256 & 48.0 & 51.1 & 55.0 & 47.8 & 51.9 & 50.8 & 76.8\% \\
    SparseVLM~{\scriptsize\textcolor{gray}{(ICML'25)}} & L1--L32 & 52.7 & 56.2 & 46.1 & 1505 & 75.1 & 62.2 & 68.2 & 51.8 & 51.1 & 53.1 & 86.7\% \\
    DART~{\scriptsize\textcolor{gray}{(EMNLP'25)}} & L2 & 55.9 & 60.6 & 53.6 & 1765 & 73.9 & 69.8 & 72.4 & 54.4 & 47.2 & 55.3 & 92.3\% \\
    HiRED~{\scriptsize\textcolor{gray}{(AAAI'25)}} & Pre & 54.6 & 60.2 & 53.0 & 1599 & 73.6 & 68.2 & 68.7 & 44.2 & - & 50.2 & 88.7\% \\
    PDrop~{\scriptsize\textcolor{gray}{(CVPR'25)}} & L8--L24 & 47.5 & 58.8 & 50.5 & 1561 & 55.9 & 69.0 & 69.2 & 50.6 & 40.0 & - & 82.7\% \\
    DivPrune~{\scriptsize\textcolor{gray}{(CVPR'25)}} & Pre & 57.6 & 59.3 & 53.7 & 1638 & 85.6 & 68.3 & 72.9 & 55.5 & 55.4 & 57.5 & 94.9\% \\
    VisionZip~{\scriptsize\textcolor{gray}{(CVPR'25)}} & Pre & 55.1 & 60.1 & 50.4 & 1690 & 77.0 & 69.0 & 72.4 & 55.5 & 54.5 & 54.8 & 92.7\% \\
    HoloV~{\scriptsize\textcolor{gray}{(NeurIPS'25)}} & Pre & 55.3 & 63.3 & 55.1 & 1715 & 80.3 & 69.5 & 72.8 & 55.4 & - & 52.8 & 94.8\% \\
    PRUNESID~{\scriptsize\textcolor{gray}{(ICLR'26)}} & Pre & 57.1 & 58.8 & - & 1733 & 83.8 & 67.8 & 73.7 & 54.2 & 56.1 & 56.9 & \textcolor{blue}{95.1}\% \\
    \rowcolor{rowblue} \textbf{AutoSelect (Ours)} & Pre & 56.8 & 62.6 & 56.6 & 1723 & 83.4 & 70.1 & 73.9 & 54.3 & 57.6 & 57.2 & \textcolor{red}{96.5}\% \\
    \bottomrule
  \end{tabular}%
  }
\end{table}

\paragraph{Results on LLaVA-1.5-7B.}
To ensure a direct comparison with prior literature, we establish LLaVA-1.5-7B as our primary testbed. In this architecture, the vision encoder processes input images at 336$\times$336 resolution, yielding a sequence of 576 visual tokens. \cref{tab:main_results} presents the comparison on LLaVA-1.5-7B across ten benchmarks.
We group methods by pruning location (Pre-LLM vs.\ In-LLM) to facilitate fair interpretation.
We evaluate three compression regimes, retaining 192, 128, and 64 tokens, corresponding to progressively more aggressive reductions from the original 576 tokens.
When retaining 192 tokens, our method achieves performance on par with the strongest baselines and is slightly below PRUNESID~\cite{fangprune} in average accuracy. However, this gap should be interpreted together with the efficiency analysis in \cref{sec:efficiency}. PRUNESID~\cite{fangprune} incurs substantially higher overhead in determining which tokens to preserve, which offsets its small performance advantage.
As the token budget decreases, the advantage of AutoSelect becomes more evident. At 128 tokens, AutoSelect surpasses all baselines in average retention. Under extreme compression at 64 tokens, AutoSelect reaches 96.5\% average retention, exceeding PRUNESID~\cite{fangprune} by 1.4\%. This suggests that the capacity-constrained formulation is particularly effective at identifying informative token subsets when the budget is limited.

\paragraph{Results on LLaVA-NEXT-7B.}
To evaluate scalability under higher visual loads, we extend our analysis to LLaVA-NEXT-7B, which raises the input resolution to $672{\times}672$ and produces $2{,}880$ visual tokens---a $5{\times}$ increase over LLaVA-1.5. This longer sequence amplifies the LLM prefill bottleneck, making effective token reduction both more critical and more challenging. We retain only 320 tokens (88.9\% reduction). As shown in \cref{tab:results_next}, AutoSelect achieves 96.1\% average performance retention, outperforming the strongest baseline (HoloV, 95.7\%) by 0.4\%, confirming that our capacity-constrained scoring mechanism generalizes well to substantially larger token pools without degradation.

\begin{table}[t]
  \centering
  \caption{
    \textbf{Results on LLaVA-NEXT-7B.}
    ``Loc.'' denotes pruning location (Pre = before LLM; L$a$--L$b$ = from LLM layer $a$ to $b$).
    ``Avg.'' is the mean ratio of each benchmark score to its upper bound.
    \textbf{Formatting:} the best Avg.\ is colored \textcolor{red}{red} and the second best is colored \textcolor{blue}{blue}; our method is highlighted in~\colorbox{rowblue}{blue}.
  }
  \label{tab:results_next}
  \scriptsize
  \setlength{\tabcolsep}{3pt}
  \resizebox{\linewidth}{!}{%
  \begin{tabular}{lcccccccccc}
    \toprule
    Method & Loc. & GQA & MMB & MMB$_{\text{CN}}$ & MME & POPE & SQA & VQA$^{\text{v2}}$ & VQA$^{\text{Text}}$ & Avg. \\
    \midrule
    \rowcolor{rowgray} \multicolumn{11}{c}{\textit{Upper Bound, 2880 Tokens (100\%)}} \\
    Vanilla & - & 64.2 & 67.4 & 60.6 & 1851 & 86.5 & 70.1 & 80.8 & 64.9 & 100\% \\
    \midrule
    \rowcolor{rowgray} \multicolumn{11}{c}{\textit{Retain 320 Tokens (88.9\% pruning)}} \\
    FastV~{\scriptsize\textcolor{gray}{(ECCV'24)}} & L2--L32 & 55.9 & 61.6 & 51.9 & 1661 & 71.7 & 62.8 & 71.9 & 55.7 & 87.6\% \\
    PDrop~{\scriptsize\textcolor{gray}{(CVPR'25)}} & L8--L24 & 56.4 & 63.4 & 56.2 & 1663 & 77.6 & 67.5 & 73.5 & 54.4 & 90.7\% \\
    DART~{\scriptsize\textcolor{gray}{(EMNLP'25)}} & L2 & 61.7 & 65.3 & 58.2 & 1710 & 84.1 & 68.4 & 79.1 & 58.7 & 95.6\% \\
    HiRED~{\scriptsize\textcolor{gray}{(AAAI'25)}} & Pre & 59.3 & 64.2 & 55.9 & 1690 & 83.3 & 66.7 & 75.7 & 58.8 & 93.4\% \\
    HoloV~{\scriptsize\textcolor{gray}{(NeurIPS'25)}} & Pre & 61.7 & 65.3 & 57.5 & 1738 & 83.9 & 68.9 & 79.5 & 58.7 & \textcolor{blue}{95.7}\% \\
    \rowcolor{rowblue} \textbf{AutoSelect (Ours)} & Pre & 62.3 & 64.7 & 57.4 & 1723 & 85.9 & 72.7 & 78.6 & 56.7 & \textcolor{red}{96.1}\% \\
    \bottomrule
  \end{tabular}%
  }
\end{table}

\paragraph{Results on Qwen2.5-VL-7B.}
We further evaluate on Qwen2.5-VL-7B~\cite{bai2025qwen25vltechnicalreport}, which differs from LLaVA in vision encoder, projector, and LLM backbone. Qwen2.5-VL processes images at native resolution, so the visual token count varies per image; we therefore report results by pruning rate rather than a fixed token budget.
As shown in \cref{tab:qwen_7b}, AutoSelect outperforms all baselines across all three pruning rates. Because the Scorer operates on per-token features without assumptions about grid layout or fixed sequence length, the same architecture and training recipe apply directly to Qwen's variable-length setting. This indicates that the method generalizes beyond the LLaVA family.

\begin{table}[t]
  \centering
  \caption{
    \textbf{Results on Qwen2.5-VL-7B.}
    ``Avg.'' is the mean ratio of each benchmark score to its upper bound.
    \textbf{Formatting:} the best Avg.\ is colored \textcolor{red}{red} and the second best is colored \textcolor{blue}{blue}; our method is highlighted in~\colorbox{rowblue}{blue}.
  }
  \label{tab:qwen_7b}
  \scriptsize
  \setlength{\tabcolsep}{4pt}
  \resizebox{0.75\linewidth}{!}{%
  \begin{tabular}{lcccccc}
    \toprule
    Method & MMB & MME & POPE & SQA & VQA$^{\text{Text}}$ & Avg. \\
    \midrule
    \rowcolor{rowgray} \multicolumn{7}{c}{\textit{Upper Bound (100\%)}} \\
    Vanilla & 82.8 & 2304 & 86.1 & 84.7 & 84.8 & 100\% \\
    \midrule
    \rowcolor{rowgray} \multicolumn{7}{c}{\textit{Token Pruning Rate = 66.7\%}} \\
    FastV~{\scriptsize\textcolor{gray}{(ECCV'24)}} & 75.7 & 2072 & 82.2 & 78.5 & 77.9 & 92.3\% \\
    HoloV~{\scriptsize\textcolor{gray}{(NeurIPS'25)}} & 78.3 & 2093 & 85.0 & 79.8 & 78.9 & \textcolor{blue}{94.3}\% \\
    \rowcolor{rowblue} \textbf{AutoSelect (Ours)} & 81.7 & 2279 & 84.9 & 86.4 & 79.0 & \textcolor{red}{98.3}\% \\
    \midrule
    \rowcolor{rowgray} \multicolumn{7}{c}{\textit{Token Pruning Rate = 77.8\%}} \\
    FastV~{\scriptsize\textcolor{gray}{(ECCV'24)}} & 74.9 & 2036 & 80.7 & 78.0 & 69.0 & 89.2\% \\
    HoloV~{\scriptsize\textcolor{gray}{(NeurIPS'25)}} & 76.5 & 2043 & 82.3 & 79.8 & 70.3 & \textcolor{blue}{90.8}\% \\
    \rowcolor{rowblue} \textbf{AutoSelect (Ours)} & 81.0 & 2218 & 82.8 & 83.7 & 76.7 & \textcolor{red}{95.9}\% \\
    \midrule
    \rowcolor{rowgray} \multicolumn{7}{c}{\textit{Token Pruning Rate = 88.9\%}} \\
    FastV~{\scriptsize\textcolor{gray}{(ECCV'24)}} & 69.2 & 1940 & 78.6 & 77.4 & 60.3 & 84.3\% \\
    HoloV~{\scriptsize\textcolor{gray}{(NeurIPS'25)}} & 72.4 & 2006 & 80.7 & 79.5 & 61.8 & \textcolor{blue}{87.0}\% \\
    \rowcolor{rowblue} \textbf{AutoSelect (Ours)} & 76.7 & 2113 & 79.8 & 82.9 & 76.7 & \textcolor{red}{93.1}\% \\
    \bottomrule
  \end{tabular}%
  }
\end{table}

\begin{figure}[t]
  \begin{minipage}[t]{0.38\linewidth}
    \centering
    \includegraphics[width=\linewidth]{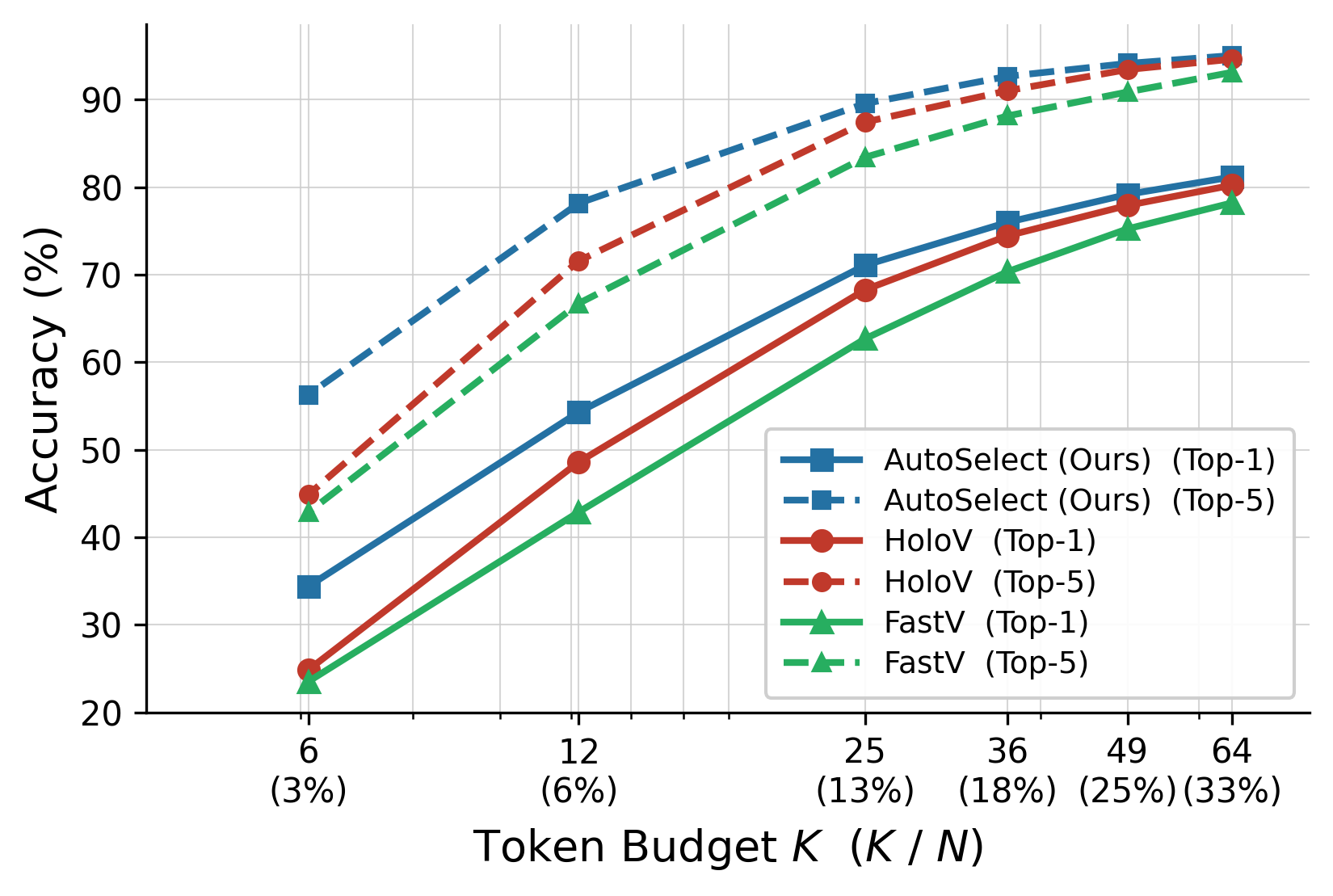}
    \captionof{figure}{
      \textbf{LLM-free classification on ImageNet-1K.}
      Each method generates a selection mask on $24{\times}24$ token grid, which is resized to $14{\times}14$ and applied to a ViT-B/16 by removing unselected patches before embedding.
    }
    \label{fig:llm_free}
  \end{minipage}
  \hfill
  \begin{minipage}[t]{0.59\linewidth}
    \centering
    \includegraphics[width=\linewidth]{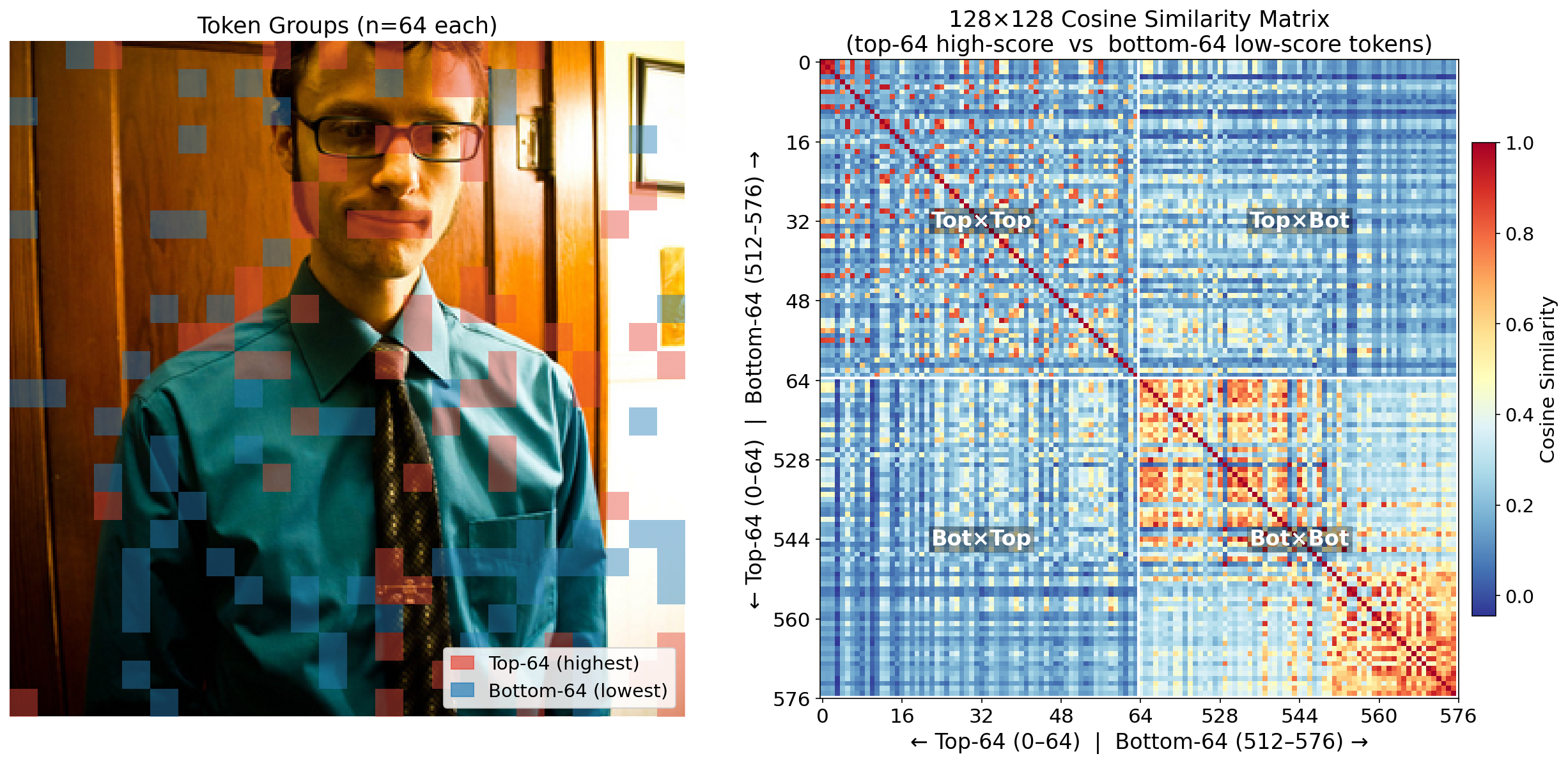}
    \captionof{figure}{
      \textbf{Token selection and pairwise similarity} ($K{=}64$). Red/blue patches denote the 64 highest-/lowest-scored tokens. The $128{\times}128$ cosine-similarity matrix is arranged as $[\text{top-}64\;|\;\text{bottom-}64]$. Retained tokens (upper-left block) are dissimilar; pruned tokens (lower-right block) are highly similar.
    }
    \label{fig:token_vis}
  \end{minipage}
\end{figure}

\subsection{LLM-Free Evaluation of Pruning Quality}
\label{sec:llm_free}
Because the Scorer operates entirely on visual features, its selection quality can be measured without the LLM.
Each method first scores all $N{=}576$ tokens on the $24{\times}24$ grid produced by CLIP ViT-L/14 and generates a binary selection mask for budget~$K$.
This mask is then resized from $24{\times}24$ to $14{\times}14$ to match the patch grid of a separately trained ViT-B/16, and unselected patches are discarded \emph{before} patch embedding, so the classifier sees only the chosen subset.
The ViT-B/16 is a MAE-pretrained model fine-tuned on ImageNet-1K.
Top-1 and Top-5 accuracy are reported for $K$ from 6 to 64.
We compare against FastV~\cite{chen2024fastv} and HoloV~\cite{holov}, representing in-LLM and pre-LLM pruning baselines respectively.

\cref{fig:llm_free} shows that AutoSelect surpasses both FastV and HoloV at every budget in both Top-1 and Top-5 accuracy.
The margin is widest under heavy pruning: at $K{=}6$ (3\% of tokens) AutoSelect leads HoloV by roughly 10 percentage points in Top-1, while FastV trails further behind.
As the budget grows the three methods gradually converge, yet AutoSelect maintains a consistent advantage at $K{=}64$.
This is expected: when only a handful of tokens survive, attention-based heuristics (FastV) and handcrafted scoring (HoloV) struggle to cover enough semantic content, whereas the learned Scorer can still place its budget on the most informative patches.
Because the experiment bypasses the LLM entirely, it confirms that the accuracy gains in \cref{sec:main_results} originate from better token selection rather than from LLM adaptation.

\subsection{Efficiency Analysis}
\label{sec:efficiency}

We profile all methods on LLaVA-1.5-7B with $336{\times}336$ inputs and $N{=}576$ visual tokens on a single NVIDIA A6000 GPU under batch size~1 with FP16 precision.
FLOPs are measured via PyTorch Profiler, and latencies are averaged over 30 forward passes after warmup.
To pinpoint savings, we decompose time-to-first-token (TTFT) into three stages: \emph{Vision Encoding}, covering the vision-tower forward pass; \emph{Pruning Module}, capturing token selection overhead; and \emph{LLM Prefill}, measuring the language-model forward pass over retained tokens.

\begin{table}[t]
  \centering
  \caption{
    \textbf{Efficiency and prefill-stage breakdown on LLaVA-1.5-7B} ($K{=}64$).
    ``Loc.'' denotes pruning location, consistent with previous tables.
    Prefill Total is time-to-first-token (TTFT).
    All latencies are in milliseconds, averaged over 30 forward passes.
  }
  \label{tab:efficiency}
  \scriptsize
  \setlength{\tabcolsep}{3.2pt}
  \resizebox{\linewidth}{!}{%
  \begin{tabular}{lccccccc}
    \toprule
    Method & Loc. & $K$ & FLOPs (T) & Vision Enc. (ms) & Pruning Module (ms) & LLM Prefill (ms) & Prefill Total (ms) \\
    \midrule
    Full (LLaVA) & ---  & 576 & 8.89 & 29.71 & 0.00 & 118.66 & 149.51 \\
    \midrule
    PruneSID & Pre & 64 & 2.13 & 31.80 & 43.39 & 39.74 & 115.84 \\
    PyramidDrop & L8--L24 & 64 & 4.89 & 29.54 & 5.02 & 70.08 & 105.90 \\
    HoloV & Pre & 64 & 2.15 & 31.96 & 2.77 & 41.48 & 77.39 \\
    \textbf{AutoSelect (Ours)} & Pre & 64 & 2.13 & 29.41 & \textbf{0.69} & 41.61 & \textbf{72.73} \\
    \bottomrule
  \end{tabular}%
  }
\end{table}

As shown in \cref{tab:efficiency}, vision encoding cost is nearly identical at approximately 30\,ms, since none prune tokens inside the encoder.
Key differences emerge subsequently.
PyramidDrop drops tokens at layers 8, 16, and 24; earlier layers attend over the full visual prefix, resulting in LLM prefill of 70.08\,ms and 4.89\,T FLOPs.
The three Pre-LLM methods, PruneSID, HoloV, and AutoSelect, reduce tokens before the LLM and share similar prefill costs of approximately 40\,ms and 2.1\,T FLOPs.
Among them, pruning overhead becomes decisive.
PruneSID's module requires 43.39\,ms, over 60$\times$ slower than AutoSelect, pushing its prefill total to 115.84\,ms, barely faster than using all tokens.
HoloV's module costs 2.77\,ms, roughly 4$\times$ that of AutoSelect.
By contrast, AutoSelect completes token selection in 0.69\,ms and achieves the lowest TTFT of 72.73\,ms.

\subsection{Visualization}
\label{sec:visualization}

\cref{fig:token_vis} visualizes the Scorer's decisions on a sample image.
In the left panel, high-score tokens (red) cover the subject's face, hands, and clothing texture, while low-score tokens (blue) land on the wooden background and plain shirt regions where neighbouring patches look alike.
The right panel makes this difference quantitative.
We compute cosine similarity between the 64 highest- and 64 lowest-scored tokens and arrange the resulting $128{\times}128$ matrix in block form.
The Top$\times$Top block (upper left) is predominantly blue: retained tokens are far apart in feature space, meaning each one carries distinct information.
The Bot$\times$Bot block (lower right) is red: pruned tokens cluster tightly, so removing them costs little unique information.
The capacity-constrained training thus pushes the Scorer to spread its budget across tokens that are far apart in feature space, rather than wasting it on near-duplicates.

\subsection{Ablation Studies}
\label{sec:ablation}

We ablate the two core design choices of our training framework, namely VP noise gating and diagonal attention in the Denoiser, on LLaVA-1.5-7B across three token budgets ($K{=}64, 128, 192$).
For each configuration, we report the average performance retention over four benchmarks (GQA, MMB, POPE, MME) as summarized in \cref{tab:ablation}.
The base configuration (VP noise gating + diagonal attention) serves as our default; each variant modifies exactly one design choice.

\begin{table}[t]
  \centering
  \caption{
    \textbf{Ablation studies} on LLaVA-1.5-7B.
    Each row reports the average performance retention (\%) over GQA, MMB, POPE, and MME across three token budgets.
    ``Base'' denotes the full AutoSelect configuration (VP noise + diagonal attention).
  }
  \label{tab:ablation}
  \small
  \setlength{\tabcolsep}{4pt}
  \resizebox{\linewidth}{!}{%
  \begin{tabular}{llccc}
    \toprule
    Configuration & Modification & $K{=}64$ & $K{=}128$ & $K{=}192$ \\
    \midrule
    Base (VP noise + diagonal)$^{\dagger}$ & -- & \textbf{95.5}\% & \textbf{97.3}\% & \textbf{98.3}\% \\
    Global attention & diagonal $\to$ global & 92.7\%{\scriptsize\,({-2.8})} & 95.0\%{\scriptsize\,({-2.3})} & 95.9\%{\scriptsize\,({-2.4})} \\
    Scale gating & VP noise $\to$ scale & 93.2\%{\scriptsize\,({-2.3})} & 94.8\%{\scriptsize\,({-2.5})} & 96.5\%{\scriptsize\,({-1.8})} \\
    \bottomrule
  \end{tabular}%
  }
\end{table}

\paragraph{VP noise gating vs.\ scale gating.}
The central mechanism of our training framework is variance-preserving noise injection, which modulates each token's information throughput via $\mathbf{x}' = \sqrt{\alpha}\,\mathbf{x} + \sqrt{1-\alpha}\,\boldsymbol{\epsilon}$.
A natural alternative is direct scale gating, $\mathbf{x}' = \alpha \cdot \mathbf{x}$, which simply attenuates low-score tokens toward zero.
\cref{tab:ablation} shows that replacing VP noise with scale gating degrades performance across all token budgets, with the gap widening under aggressive pruning ($K{=}64$: 95.5\% vs.\ 93.2\%).
Scale gating reduces the magnitude of low-importance tokens but preserves their directional information intact, allowing the downstream LLM to partially recover the attenuated content through its layernorm and attention mechanisms.
VP noise gating, by contrast, actively corrupts the representational content of low-score tokens with isotropic noise, creating a hard capacity constraint during training: the only way for the system to preserve task-relevant information is to assign high importance scores to the corresponding tokens.
This creates a stronger learning signal for the Scorer, resulting in more discriminative importance allocation.

\paragraph{Diagonal vs.\ global attention in the Denoiser.}
Replacing diagonal attention with global self-attention causes the largest degradation across all budgets (\cref{tab:ablation}), confirming the information-leakage hypothesis.
With global attention, heavily noised tokens can attend to high-score tokens and recover information that should have been suppressed, effectively circumventing the capacity constraint imposed by the noise gating stage.
Diagonal attention enforces strict per-token independence, ensuring that each token's information throughput is determined solely by its importance score.
This design is essential for maintaining the integrity of the capacity-constrained formulation during training.

\begin{figure}[t]
  \centering
  \begin{subfigure}{0.52\linewidth}
    \includegraphics[width=\linewidth]{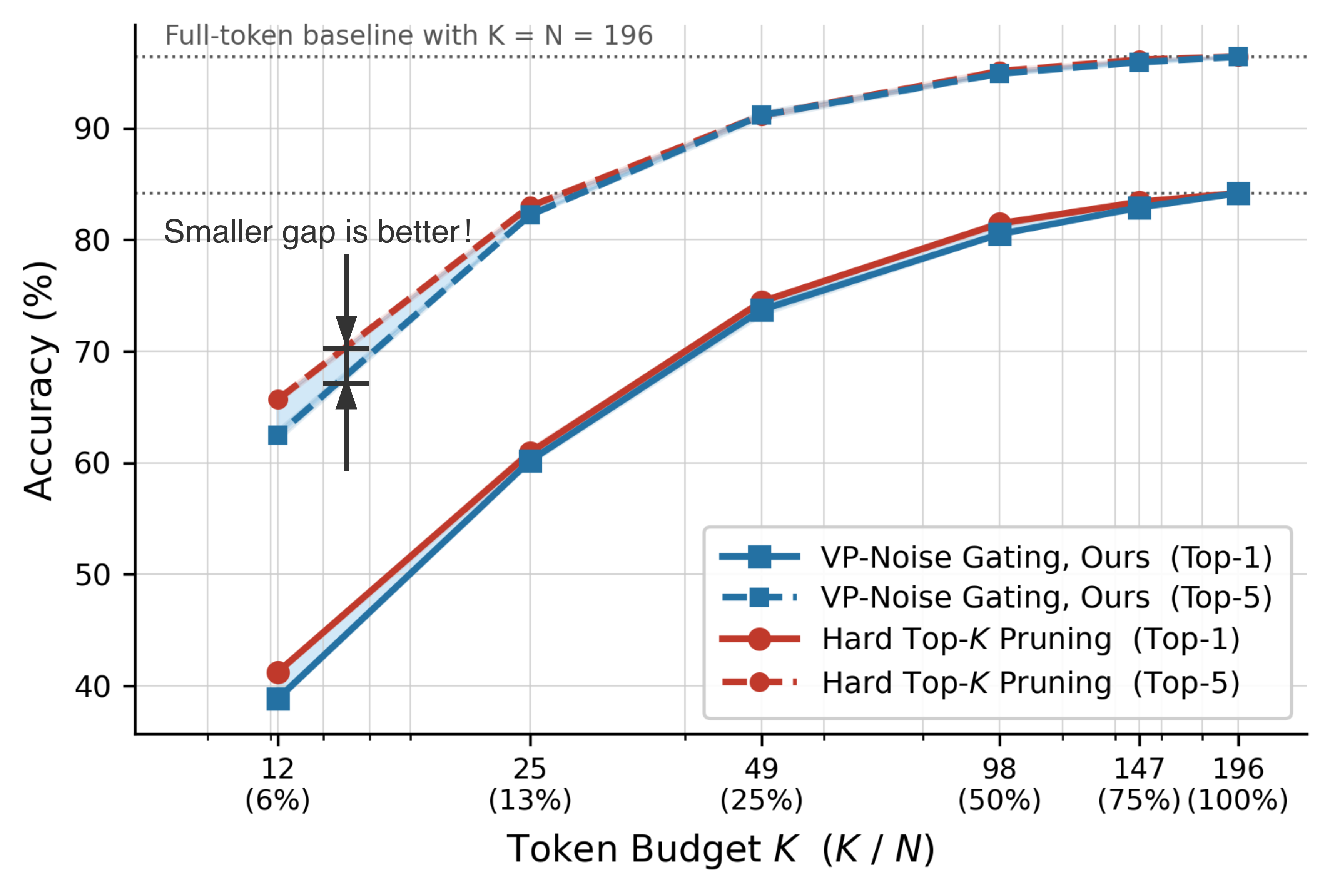}
    \caption{Top-1 (solid) and Top-5 (dashed) accuracy on ImageNet-1K validation as a function of token budget~$K$. Shaded regions indicate the gap between the two strategies. Dotted lines mark the full-token baseline ($K{=}N$).}
    \label{fig:vp_validation_quant}
  \end{subfigure}
  \hfill
  \begin{subfigure}{0.45\linewidth}
    \includegraphics[width=\linewidth]{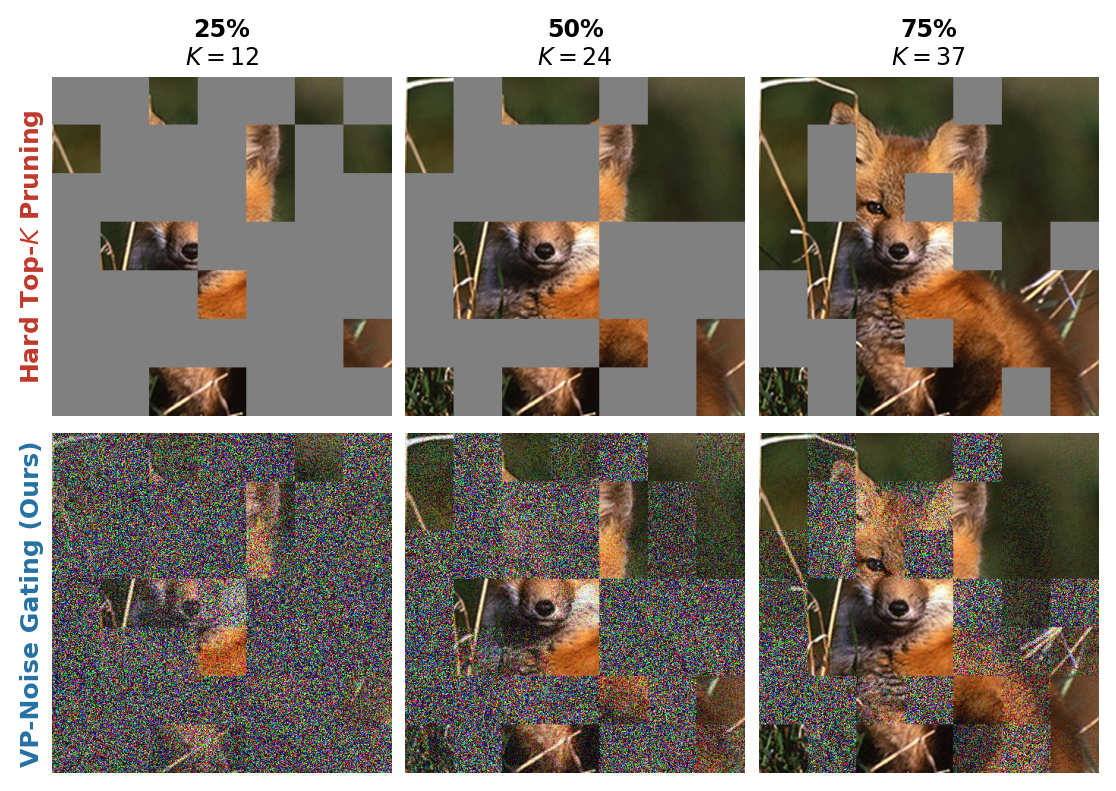}
    \caption{Visualization under three retention levels with $N{=}49$ tokens ($K\in\{12,24,37\}$, corresponding to $\sim$25\%, 50\%, and 75\% retention).
\textit{Top:} hard Top-$K$ masking keeps the $K$ selected patches while graying out the rest.
\textit{Bottom:} VP-noise gating retains all positions but injects score-proportional noise, producing a soft selection effect.}
    \label{fig:vp_validation_qual}
  \end{subfigure}
  \caption{
    \textbf{Validation of VP-noise gating as a differentiable proxy for hard Top-$K$ pruning.}
    Both strategies are applied after patch embedding and before the vision encoder, matching the insertion point used by AutoSelect.
    Both use identical random score maps to control for scoring quality.
    Quantitatively~(a) and qualitatively~(b), the two mechanisms produce equivalent information restriction across all budget levels, justifying VP-noise gating as a continuous, gradient-friendly training surrogate.
  }
  \label{fig:vp_validation}
\end{figure}

\paragraph{VP noise gating vs.\ hard Top-$K$ pruning.}
We empirically verify that VP noise gating imposes information constraints similar to discrete removal.
Using a random scorer to isolate the gating mechanism, \cref{fig:vp_validation_quant} shows that both strategies produce nearly identical accuracy degradation on ImageNet-1K across token budgets.
The narrow shaded gap in \cref{fig:vp_validation_quant} confirms that VP noise gating closely matches the information restriction imposed by hard Top-$K$, and \cref{fig:vp_validation_qual} provides a visual comparison at matched retention levels.

\section{Conclusion}
\label{sec:conclusion}

We presented AutoSelect, a framework that recasts visual token pruning as capacity-constrained representation learning.
By modulating per-token information throughput via variance-preserving noise rather than discarding tokens outright, the framework turns discrete pruning into a continuous optimization problem trained end-to-end with the standard next-token prediction loss as the sole objective, requiring no auxiliary losses.
A diagonal-attention Denoiser prevents information leakage across tokens during training and is removed at inference, adding zero overhead.
On LLaVA-1.5-7B, AutoSelect retains 96.5\% of full-model accuracy at 88.9\% token pruning with only 0.69\,ms selection overhead, and generalizes without architecture-specific modification to LLaVA-NEXT and Qwen2.5-VL, consistently outperforming existing methods across all evaluated settings.
These results indicate that learned capacity allocation can replace heuristic pruning criteria: given a fixed bandwidth budget and a differentiable relaxation, the model discovers which visual tokens carry task-relevant information.

\bibliographystyle{unsrt}
\bibliography{references}

@String(CVPR  = {IEEE Conf. Comput. Vis. Pattern Recog.})

@String(ICCV  = {Int. Conf. Comput. Vis.})

@String(ECCV  = {Eur. Conf. Comput. Vis.})

@String(NeurIPS = {Adv. Neural Inform. Process. Syst.})

@String(ICML  = {Int. Conf. Mach. Learn.})

@String(ICLR  = {Int. Conf. Learn. Represent.})

@String(AAAI  = {AAAI})

@String(CVPR  = {CVPR})

@String(ICCV  = {ICCV})

@String(ECCV  = {ECCV})

@String(NeurIPS = {NeurIPS})

@String(ICML  = {ICML})

@String(ICLR  = {ICLR})

@inproceedings{
    yang2025one,
    title={One Leaf Reveals the Season: Occlusion-Based Contrastive Learning with Semantic-Aware Views for Efficient Visual Representation},
    author={Yang, Xiaoyu and Xu, Lijian and Li, Hongsheng and Zhang, Shaoting},
    booktitle={Forty-second International Conference on Machine Learning},
    year={2025}
}

@article{young2026fewer,
  title={Fewer Tokens, Greater Scaling: Self-Adaptive Visual Bases for Efficient and Expansive Representation Learning},
  author={Young, Shawn and Zeng, Xingyu and Xu, Lijian},
  journal={arXiv preprint arXiv:2511.19515},
  year={2026}
}

@article{young2026scalar,
  title={SCALAR: Spatial-Concept Alignment for Robust Vision in Harsh Open World},
  author={Yang, Xiaoyu and Xu, Lijian and Zeng, Xingyu and Wang, Xiaosong and Li, Hongsheng and Zhang, Shaoting},
  journal={Pattern Recognition},
  year={2026}
}

@article{xu2023learning,
  title={Learning a multi-task transformer via unified and customized instruction tuning for chest radiograph interpretation},
  author={Xu, Lijian and Ni, Ziyu and Liu, Xinglong and Wang, Xiaosong and Li, Hongsheng and Zhang, Shaoting},
  journal={arXiv preprint arXiv:2311.01092},
  year={2023}
}

@article{yang2024segmentation,
  title={Segmentation and vascular vectorization for coronary artery by geometry-based cascaded neural network},
  author={Yang, Xiaoyu and Xu, Lijian and Yu, Simon and Xia, Qing and Li, Hongsheng and Zhang, Shaoting},
  journal={IEEE Transactions on Medical Imaging},
  year={2024},
  publisher={IEEE}
}

@inproceedings{li2023blip2,
  title={BLIP-2: Bootstrapping Language-Image Pre-training with Frozen Image Encoders and Large Language Models},
  author={Li, Junnan and Li, Dongxu and Savarese, Silvio and Hoi, Steven},
  booktitle={Proceedings of the 40th International Conference on Machine Learning (ICML)},
  pages={19730--19742},
  year={2023},
  organization={PMLR}
}

@inproceedings{chen2024fastv,
  title={An Image is Worth 1/2 Tokens After Layer 2: Plug-and-Play Inference Acceleration for Large Vision-Language Models},
  author={Chen, Liang and Zhao, Haozhe and Liu, Tianyu and Bai, Shuang and Lin, Junyang and Zhou, Chang and Chang, Baobao},
  booktitle={Proceedings of the European Conference on Computer Vision (ECCV)},
  year={2024}
}

@inproceedings{yang2025visionzip,
  title={VisionZip: Longer is Better but Not Necessary in Vision Language Models},
  author={Yang, Senqiao and Chen, Yukang and Tian, Zhuotao and Wang, Chengyao and Li, Jingyao and Yu, Bei and Jia, Jiaya},
  booktitle={Proceedings of the IEEE/CVF Conference on Computer Vision and Pattern Recognition (CVPR)},
  pages={19792--19802},
  year={2025}
}

@inproceedings{liu2023visual_instruction_tuning,
  title={{LLaVA}: Large Language and Vision Assistant for Visual Instruction Tuning},
  author={Liu, Haotian and others},
  booktitle={Advances in Neural Information Processing Systems (NeurIPS) 36},
  year={2023}
}

@inproceedings{dai2023instructblip,
  title={InstructBLIP: Towards General-purpose Vision-Language Models with Instruction Tuning},
  author={Dai, Wenliang and Li, Junnan and Li, Dongxu and Tiong, Anthony Meng Huat and Zhao, Junqi and Wang, Weisheng and Li, Boyang and Fung, Pascale and Hoi, Steven},
  booktitle={Advances in Neural Information Processing Systems (NeurIPS) 36},
  year={2023}
}

@inproceedings{instructblip,
    title = {Instruct{BLIP}: Towards General-purpose Vision-Language Models with Instruction Tuning},
    author = {Dai, Wenliang and others},
    booktitle = NeurIPS,
    year = {2023}
}

@inproceedings{yang2025topv,
  title={TopV: Compatible Token Pruning with Inference Time Optimization for Fast and Low-Memory Multimodal Vision Language Model},
  author={Yang, Cheng and Sui, Yang and Xiao, Jinqi and Huang, Lingyi and Gong, Yu and Li, Chendi and Yan, Jinghua and Bai, Yu and Sadayappan, Ponnuswamy and Hu, Xia and Yuan, Bo},
  booktitle={Proceedings of the IEEE/CVF Conference on Computer Vision and Pattern Recognition (CVPR)},
  pages={19803--19813},
  year={2025}
}

@inproceedings{zhao2025_gsearch,
  title={Accelerating Multimodal Large Language Models by Searching Optimal Vision Token Reduction},
  author={Zhao, Shiyu and Wang, Zhenting and Juefei-Xu, Felix and Xia, Xide and Liu, Miao and Wang, Xiaofang and Liang, Mingfu and Zhang, Ning and Metaxas, Dimitris N. and Yu, Licheng},
  booktitle={Proceedings of the IEEE/CVF Conference on Computer Vision and Pattern Recognition (CVPR)},
  pages={29869--29879},
  year={2025}
}

@inproceedings{huang2024ivtp,
  title={IVTP: Instruction-guided Visual Token Pruning for Large Vision-Language Models},
  author={Huang, Kai and Zou, Hao and Xi, Ye and Wang, BoChen and Xie, Zhen and Yu, Liang},
  booktitle={Proceedings of the European Conference on Computer Vision (ECCV)},
  pages={214--230},
  year={2024}
}

@inproceedings{cao2024madtp,
  title={MADTP: Multimodal Alignment-Guided Dynamic Token Pruning for Accelerating Vision-Language Transformer},
  author={Cao, Jianjian and Ye, Peng and Li, Shengze and Yu, Chong and Tang, Yansong and Lu, Jiwen and Chen, Tao},
  booktitle={Proceedings of the IEEE/CVF Conference on Computer Vision and Pattern Recognition (CVPR)},
  pages={15710--15719},
  year={2024}
}

@inproceedings{ye2025_atp_llava,
  title={ATP-LLaVA: Adaptive Token Pruning for Large Vision Language Models},
  author={Ye, Xubing and Gan, Yukang and Ge, Yixiao and Zhang, Xiao-Ping and Tang, Yansong},
  booktitle={Proceedings of the IEEE/CVF Conference on Computer Vision and Pattern Recognition (CVPR)},
  pages={24972--24982},
  year={2025}
}

@misc{flamingo,
    title = {Flamingo: a Visual Language Model for Few-Shot Learning},
    author = {Alayrac, Jean-Baptiste and others},
    year = {2022},
    booktitle = NeurIPS,
}

@inproceedings{eve,
    title = {Unveiling Encoder-Free Vision-Language Models},
    author = {Zhang, Zhiyuan and others},
    booktitle = NeurIPS,
    year = {2024}
}

@misc{llava-next-interleave,
    title = {{LLaVA}-NeXT-Interleave: Tackling Multi-image, Video, and 3D in Large Multimodal Models},
    author = {Li, Xin and others},
    year = {2025},
    note = {ICLR}
}

@misc{llava-onevision,
    title = {{LLaVA}-OneVision: Easy Visual Task Transfer},
    author = {Li, Xin and others},
    year = {2024},
    note = {arXiv preprint}
}

@misc{qwen2-vl,
    title = {{Qwen2}-{VL}: Enhancing Vision-Language Model's Perception of the World at Any Resolution},
    author = {Qwen Team},
    year = {2024},
    note = {arXiv preprint}
}

@misc{mra,
    title = {Feast Your Eyes: Mixture-of-Resolution Adaptation for Multimodal Large Language Models},
    author = {Yu, Jiahui and others},
    year = {2026},
    note = {ICLR}
}

@inproceedings{flexattention,
    title = {FlexAttention for Efficient High-Resolution Vision-Language Models},
    author = {Chen, Yuxin and others},
    booktitle = ECCV,
    year = {2024}
}

@inproceedings{fastvlm,
    title = {Fast{VLM}: Efficient Vision Encoding for Vision Language Models},
    author = {Wang, Zekun and others},
    booktitle = CVPR,
    year = {2025}
}

@inproceedings{swin,
    title = {Swin Transformer: Hierarchical Vision Transformer using Shifted Windows},
    author = {Liu, Ze and Lin, Yutong and Cao, Yue and Hu, Han and Wei, Yixuan and Zhang, Zheng and Lin, Stephen and Guo, Baining},
    booktitle = ICCV,
    year = {2021}
}

@article{flashattention,
  title={Flashattention: Fast and memory-efficient exact attention with io-awareness},
  author={Dao, Tri and Fu, Dan and Ermon, Stefano and Rudra, Atri and R{\'e}, Christopher},
  journal={Advances in neural information processing systems},
  volume={35},
  pages={16344--16359},
  year={2022}
}

@article{flashattention2,
  title={Flashattention-2: Faster attention with better parallelism and work partitioning},
  author={Dao, Tri},
  journal={arXiv preprint arXiv:2307.08691},
  year={2023}
}

@article{flashattention3,
  title={Flashattention-3: Fast and accurate attention with asynchrony and low-precision},
  author={Shah, Jay and Bikshandi, Ganesh and Zhang, Ying and Thakkar, Vijay and Ramani, Pradeep and Dao, Tri},
  journal={Advances in Neural Information Processing Systems},
  volume={37},
  pages={68658--68685},
  year={2024}
}

@inproceedings{tome,
    title = {Token Merging: Your {ViT} But Faster},
    author = {Bolya, Daniel and Fu, Cheng-Yang and Dai, Xiaodong and Zhang, Peize and Hoffman, Judy},
    booktitle = ICLR,
    year = {2023}
}

@inproceedings{dynamic-token-pruning,
    title = {Dynamic Token Pruning in Plain Vision Transformers for Semantic Segmentation},
    author = {Tang, Shixiang and others},
    booktitle = ICCV,
    year = {2023}
}

@inproceedings{zerotprune,
    title = {Zero-{TPrune}: Zero-Shot Token Pruning through Leveraging of the Attention Graph},
    author = {Zhang, Rui and others},
    booktitle = CVPR,
    year = {2024}
}

@inproceedings{prumerge,
    title = {{LLaVA}-{PruMerge}: Adaptive Token Reduction for Efficient Large Multimodal Models},
    author = {Chen, Jiacheng and others},
    booktitle={Proceedings of the IEEE/CVF International Conference on Computer Vision},
  pages={22857--22867},
  year={2025}
}

@inproceedings{vtw,
  title={Boosting multimodal large language models with visual tokens withdrawal for rapid inference},
  author={Lin, Zhihang and Lin, Mingbao and Lin, Luxi and Ji, Rongrong},
  booktitle={Proceedings of the AAAI Conference on Artificial Intelligence},
  volume={39},
  number={5},
  pages={5334--5342},
  year={2025}
}

@inproceedings{fitprune,
    title = {Fit and Prune: Fast and Training-free Visual Token Pruning for Multi-modal Large Language Models},
    author = {Li, Yichen and others},
    booktitle = AAAI,
    year = {2025}
}

@misc{fastervlm,
    title = {[{CLS}] Attention is All You Need for Training-Free Visual Token Pruning: Make {VLM} Inference Faster},
    author = {Zhang, Haoran and others},
    year = {2024},
    note = {arXiv preprint}
}

@inproceedings{divprune,
    title = {DivPrune: Diversity-based Visual Token Pruning for Large Multimodal Models},
    author = {Liu, Yufei and others},
    booktitle = CVPR,
    year = {2025}
}

@inproceedings{balanced-token-pruning,
    title = {Balanced Token Pruning: Accelerating Vision Language Models Beyond Local Optimization},
    author = {Sun, Qing and others},
    booktitle = NeurIPS,
    year = {2025}
}

@inproceedings{dart,
  title={Stop Looking for “Important Tokens” in Multimodal Language Models: Duplication Matters More},
  author={Wen, Zichen and Gao, Yifeng and Wang, Shaobo and Zhang, Junyuan and Zhang, Qintong and Li, Weijia and He, Conghui and Zhang, Linfeng},
  booktitle={Proceedings of the 2025 Conference on Empirical Methods in Natural Language Processing},
  pages={9972--9991},
  year={2025}
}

@misc{token-pruning-right-problem,
    title = {Token Pruning in Multimodal Large Language Models: Are We Solving the Right Problem?},
    author = {Guo, Peng and others},
    year = {2025},
    note = {Findings of ACL}
}

@inproceedings{feather-throttle,
    title = {Feather the Throttle: Revisiting Visual Token Pruning for Vision-Language Model Acceleration},
    author = {Xu, Ming and others},
    booktitle = ICCV,
    year = {2025}
}

@article{sparsevlm,
  title={Sparsevlm: Visual token sparsification for efficient vision-language model inference},
  author={Zhang, Yuan and Fan, Chun-Kai and Ma, Junpeng and Zheng, Wenzhao and Huang, Tao and Cheng, Kuan and Gudovskiy, Denis and Okuno, Tomoyuki and Nakata, Yohei and Keutzer, Kurt and others},
  journal={ICML},
  year={2025}
}

@inproceedings{fangprune,
  title={Prune Redundancy, Preserve Essence: Vision Token Compression in VLMs via Synergistic Importance-Diversity},
  author={Fang, Zhengyao and Lyu, Pengyuan and Zhang, Chengquan and Lu, Guangming and Yu, Jun and Pei, Wenjie},
  booktitle={The Fourteenth International Conference on Learning Representations},
  year={2025}
}

@inproceedings{hired,
  title={Hired: Attention-guided token dropping for efficient inference of high-resolution vision-language models},
  author={Arif, Kazi Hasan Ibn and Yoon, JinYi and Nikolopoulos, Dimitrios S and Vandierendonck, Hans and John, Deepu and Ji, Bo},
  booktitle={Proceedings of the AAAI Conference on Artificial Intelligence},
  volume={39},
  number={2},
  pages={1773--1781},
  year={2025}
}

@article{holov,
  title={Don't Just Chase" Highlighted Tokens" in MLLMs: Revisiting Visual Holistic Context Retention},
  author={Zou, Xin and Lu, Di and Wang, Yizhou and Yan, Yibo and Lyu, Yuanhuiyi and Zheng, Xu and Zhang, Linfeng and Hu, Xuming},
  journal={NeurIPS},
  year={2025}
}

@article{pyramiddrop,
  title={Pyramiddrop: Accelerating your large vision-language models via pyramid visual redundancy reduction},
  author={Xing, Long and Huang, Qidong and Dong, Xiaoyi and Lu, Jiajie and Zhang, Pan and Zang, Yuhang and Cao, Yuhang and He, Conghui and Wang, Jiaqi and Wu, Feng and others},
  journal={CVPR},
  year={2025}
}

@article{shen2025vlm,
  title={Vlm-r1: A stable and generalizable r1-style large vision-language model},
  author={Shen, Haozhan and Liu, Peng and Li, Jingcheng and Fang, Chunxin and Ma, Yibo and Liao, Jiajia and Shen, Qiaoli and Zhang, Zilun and Zhao, Kangjia and Zhang, Qianqian and others},
  journal={arXiv preprint arXiv:2504.07615},
  year={2025}
}

@article{kim2024image,
  title={An image grid can be worth a video: Zero-shot video question answering using a vlm},
  author={Kim, Wonkyun and Choi, Changin and Lee, Wonseok and Rhee, Wonjong},
  journal={IEEE Access},
  volume={12},
  pages={193057--193075},
  year={2024},
  publisher={IEEE}
}

@inproceedings{xu2021vlm,
  title={Vlm: Task-agnostic video-language model pre-training for video understanding},
  author={Xu, Hu and Ghosh, Gargi and Huang, Po-Yao and Arora, Prahal and Aminzadeh, Masoumeh and Feichtenhofer, Christoph and Metze, Florian and Zettlemoyer, Luke},
  booktitle={Findings of the Association for Computational Linguistics: ACL-IJCNLP 2021},
  pages={4227--4239},
  year={2021}
}

@article{zeng2023x,
  title={X$^{2}$-VLM: All-in-One Pre-Trained Model for Vision-Language Tasks},
  author={Zeng, Yan and Zhang, Xinsong and Li, Hang and Wang, Jiawei and Zhang, Jipeng and Zhou, Wangchunshu},
  journal={IEEE transactions on pattern analysis and machine intelligence},
  volume={46},
  number={5},
  pages={3156--3168},
  year={2023},
  publisher={IEEE}
}

@inproceedings{cao2023pumer,
  title={PuMer: Pruning and merging tokens for efficient vision language models},
  author={Cao, Qingqing and Paranjape, Bhargavi and Hajishirzi, Hannaneh},
  booktitle={Proceedings of the 61st Annual Meeting of the Association for Computational Linguistics (Volume 1: Long Papers)},
  pages={12890--12903},
  year={2023}
}

@article{li2025mini,
  title={Mini-gemini: Mining the potential of multi-modality vision language models},
  author={Li, Yanwei and Zhang, Yuechen and Wang, Chengyao and Zhong, Zhisheng and Chen, Yixin and Chu, Ruihang and Liu, Shaoteng and Jia, Jiaya},
  journal={IEEE Transactions on Pattern Analysis and Machine Intelligence},
  year={2025},
  publisher={IEEE}
}

@misc{bai2025qwen25vltechnicalreport,
  title={Qwen2.5-VL Technical Report},
  author={Shuai Bai and Keqin Chen and Xuejing Liu and Jialin Wang and Wenbin Ge and Sibo Song and Kai Dang and Peng Wang and Shijie Wang and Jun Tang and Humen Zhong and Yuanzhi Zhu and Mingkun Yang and Zhaohai Li and Jianqiang Wan and Pengfei Wang and Wei Ding and Zheren Fu and Yiheng Xu and Jiabo Ye and Xi Zhang and Tianbao Xie and Zesen Cheng and Hang Zhang and Zhibo Yang and Haiyang Xu and Junyang Lin},
  year={2025},
  eprint={2502.13923},
  archivePrefix={arXiv},
  primaryClass={cs.CV},
  url={https://arxiv.org/abs/2502.13923},
}

@inproceedings{cao2025survey,
  title={A Survey on Visual Token Compression for Efficient Vision-Language Models},
  author={Cao, Weidong and Feng, Zeqin and Sun, Fangwei and Liu, Danjun and Xie, Yongqiang and Li, Zhongbo},
  booktitle={2025 5th International Conference on Advanced Algorithms and Neural Networks (AANN)},
  pages={735--741},
  year={2025},
  organization={IEEE}
}

@online{softtopk-su,
  title  = {Softmax Hou Zhuan: Xunzhao Top-K de Guanghua Jinsi},
  author = {Su, Jianlin},
  year={2024},
  month={Sep},
  url={https://kexue.fm/archives/10373},
}

@inproceedings{pope,
  title={Evaluating object hallucination in large vision-language models},
  author={Li, Yifan and Du, Yifan and Zhou, Kun and Wang, Jinpeng and Zhao, Wayne Xin and Wen, Ji-Rong},
  booktitle={Proceedings of the 2023 conference on empirical methods in natural language processing},
  pages={292--305},
  year={2023}
}

@inproceedings{hudson2019gqa,
  title={Gqa: A new dataset for real-world visual reasoning and compositional question answering},
  author={Hudson, Drew A and Manning, Christopher D},
  booktitle={Proceedings of the IEEE/CVF conference on computer vision and pattern recognition},
  pages={6700--6709},
  year={2019}
}

@inproceedings{liu2024mmbench,
  title={Mmbench: Is your multi-modal model an all-around player?},
  author={Liu, Yuan and Duan, Haodong and Zhang, Yuanhan and Li, Bo and Zhang, Songyang and Zhao, Wangbo and Yuan, Yike and Wang, Jiaqi and He, Conghui and Liu, Ziwei and others},
  booktitle={European conference on computer vision},
  pages={216--233},
  year={2024},
  organization={Springer}
}

@article{fu2023mme,
  title={Mme: A comprehensive evaluation benchmark for multimodal large language models},
  author={Fu, Chaoyou and Chen, Peixian and Shen, Yunhang and Qin, Yulei and Zhang, Mengdan and Lin, Xu and Yang, Jinrui and Zheng, Xiawu and Li, Ke and Sun, Xing and others},
  journal={NeurIPS},
  year={2025}
}

@article{lu2022learn,
  title={Learn to explain: Multimodal reasoning via thought chains for science question answering},
  author={Lu, Pan and Mishra, Swaroop and Xia, Tanglin and Qiu, Liang and Chang, Kai-Wei and Zhu, Song-Chun and Tafjord, Oyvind and Clark, Peter and Kalyan, Ashwin},
  journal={Advances in neural information processing systems},
  volume={35},
  pages={2507--2521},
  year={2022}
}

@inproceedings{goyal2017making,
  title={Making the v in vqa matter: Elevating the role of image understanding in visual question answering},
  author={Goyal, Yash and Khot, Tejas and Summers-Stay, Douglas and Batra, Dhruv and Parikh, Devi},
  booktitle={Proceedings of the IEEE conference on computer vision and pattern recognition},
  pages={6904--6913},
  year={2017}
}

@inproceedings{singh2019towards,
  title={Towards vqa models that can read},
  author={Singh, Amanpreet and Natarajan, Vivek and Shah, Meet and Jiang, Yu and Chen, Xinlei and Batra, Dhruv and Parikh, Devi and Rohrbach, Marcus},
  booktitle={Proceedings of the IEEE/CVF conference on computer vision and pattern recognition},
  pages={8317--8326},
  year={2019}
}

@inproceedings{bigham2010vizwiz,
  title={Vizwiz: nearly real-time answers to visual questions},
  author={Bigham, Jeffrey P and Jayant, Chandrika and Ji, Hanjie and Little, Greg and Miller, Andrew and Miller, Robert C and Miller, Robin and Tatarowicz, Aubrey and White, Brandyn and White, Samual and others},
  booktitle={Proceedings of the 23nd annual ACM symposium on User interface software and technology},
  pages={333--342},
  year={2010}
}

@inproceedings{deng2009imagenet,
  title={Imagenet: A large-scale hierarchical image database},
  author={Deng, Jia and Dong, Wei and Socher, Richard and Li, Li-Jia and Li, Kai and Fei-Fei, Li},
  booktitle={2009 IEEE conference on computer vision and pattern recognition},
  pages={248--255},
  year={2009},
  organization={IEEE}
}

@dataset{visual_layer_imagenet1k_vl_enriched,
  title={ImageNet-1K-VL-Enriched},
  author={{Visual Layer}},
  year={2023},
  url={https://huggingface.co/datasets/visual-layer/imagenet-1k-vl-enriched},
  note={HuggingFace dataset}
}

@misc{liu2024llavanext,
  title={LLaVA-NeXT: Improved reasoning, OCR, and world knowledge},
  url={https://llava-vl.github.io/blog/2024-01-30-llava-next/},
  author={Liu, Haotian and Li, Chunyuan and Li, Yuheng and Li, Bo and Zhang, Yuanhan and Shen, Sheng and Lee, Yong Jae},
  month={January},
  year={2024}
}

@article{zeng2025glimpse,
  title={A Glimpse to Compress: Dynamic Visual Token Pruning for Large Vision-Language Models},
  author={Zeng, Quan-Sheng and Li, Yunheng and Wang, Qilong and Jiang, Peng-Tao and Wu, Zuxuan and Cheng, Ming-Ming and Hou, Qibin},
  journal={arXiv preprint arXiv:2508.01548},
  year={2025}
}

@inproceedings{zhang2025pmod,
  title={p-MoD: Building Mixture-of-Depths MLLMs via Progressive Ratio Decay},
  author={Zhang, Jun and Meng, Desen and Zhang, Zhengming and Huang, Zhenpeng and Wu, Tao and Wang, Limin},
  booktitle={Proceedings of the IEEE/CVF International Conference on Computer Vision},
  pages={3705--3715},
  year={2025}
}

@inproceedings{li2023seed,
  title={Seed-bench: Benchmarking multimodal large language models},
  author={Li, Bohao and Ge, Yuying and Ge, Yixiao and Wang, Guangzhi and Wang, Rui and Zhang, Ruimao and Shan, Ying},
  booktitle={Proceedings of the IEEE/CVF Conference on Computer Vision and Pattern Recognition},
  pages={13299--13308},
  year={2024}
}

@article{chen2026tc,
  title={TC-SSA: Token Compression via Semantic Slot Aggregation for Gigapixel Pathology Reasoning},
  author={Chen, Zhuo and Young, Shawn and Xu, Lijian},
  journal={arXiv preprint arXiv:2603.01143},
  year={2026}
}

@article{gao2026zerosense,
  title={ZeroSense: How Vision matters in Long Context Compression},
  author={Gao, Yonghan and Chen, Zehong and Xu, Lijian and Chen, Jingzhi and Guan, Jingwei and Zeng, Xingyu},
  journal={arXiv preprint arXiv:2603.7337970},
  year={2026}
}

@article{wu2026towards,
  title={Towards efficient multimodal large language models of Gigapixel Pathology: A Survey on Token Compression},
  author={Wu, Peihang and Xu, Lijian},
  journal={arXiv preprint},
  year={2026}
}

@article{xu2024medvilam,
  title={MedViLaM: A multimodal large language model with advanced generalizability and explainability for medical data understanding and generation},
  author={Xu, Lijian and Sun, Hao and Ni, Ziyu and Li, Hongsheng and Zhang, Shaoting},
  journal={arXiv preprint arXiv:2409.19684},
  year={2024}
}

@article{xu2024foundation,
  title={A foundation model for generalizable disease diagnosis in chest X-ray images},
  author={Xu, Lijian and Ni, Ziyu and Sun, Hao and Li, Hongsheng and Zhang, Shaoting},
  journal={arXiv preprint arXiv:2410.08861},
  year={2024}
}

@inproceedings{yang2023geometry,
  title={Geometry-based end-to-end segmentation of coronary artery in computed tomography angiography},
  author={Yang, Xiaoyu and Xu, Lijian and Yu, Simon and Xia, Qing and Li, Hongsheng and Zhang, Shaoting},
  booktitle={International Workshop on Trustworthy Machine Learning for Healthcare},
  pages={190--196},
  year={2023},
  organization={Springer}
}

@article{feng2026efficient,
  title     = {Efficient Chest X-ray Representation Learning via Semantic-Partitioned Contrastive Learning},
  author    = {Feng, Wangyu and Young, Shawn and Xu, Lijian},
  journal={arXiv preprint arXiv:2603.7338028},
  year={2026}
}

\end{document}